# Unified Complex-valued Neural Network: A Magnitude-Phase Computational Model for Event-Driven Neuromorphic Learning

**Reza Ahmadvand, Sarah Safura Sharif, Yaser Mike Banad**
Department of Electrical and Computer Engineering, University of Oklahoma, Oklahoma, United States

**Email:**
iamrezaahmadvand1@ou.edu, s.sh@ou.edu, bana@ou.edu

**Abstract**
Artificial neural networks (ANN) provide accurate continuous-valued representation, whereas spiking neural networks (SNN) offer event-driven temporal processing, yet both paradigms face limitations when value encoding and timing dynamics must be learned within a single computational structure. This paper introduces a network based on Unified Complex-valued Neuron (UCN), a new neural computational model that integrates continuous activation and phase-driven event generation through an asymmetric complex-valued state. In the UCN, magnitude encodes signal strength while phase governs intrinsic temporal evolution and valued spike emission. A foundational training framework combining backpropagation (BP) and backpropagation through time (BPTT) is first developed to optimize magnitude and phase pathways in a unified way. To reduce computational complexity, an event-driven adaptive phase learning (EAPL) rule is then introduced as a more efficient alternative. The proposed model is evaluated through object tracking and Lorenz attractor learning. Results demonstrate that UCN-based Network (UCNN) provides accurate, stable, and interpretable spatiotemporal learning while preserving sparse event-driven computation for neuromorphic and edge-AI applications..

## 1. Introduction

Artificial neural networks (ANNs) have established themselves as the preeminent computational model for modern machine learning, delivering highly accurate, continuous-valued function approximation [1]. However, conventional ANNs are fundamentally static and lack intrinsic event timing, which heavily influences performance comparisons when evaluated against dynamic systems [2]. This structural limitation creates a significant computational bottleneck in real-time edge-AI and neuromorphic perception systems, where processing dynamic spatiotemporal information efficiently is paramount [3].

Spiking neural networks (SNNs) address these inefficiencies by introducing asynchronous, event-driven computation, an architecture that has driven the development of neuromorphic processors such as Loihi, novel algorithmic opportunities, and large-scale computing platforms [4-6]. To bridge the gap between continuous and spiking paradigms, hybrid architectures such as the Tianjic chip have been developed [7]. Concurrently, event-based vision and computational event-driven sensors have demonstrated the critical value of asynchronous temporal information for low-latency perception [8, 9]. Despite hardware advances in biologically inspired artificial neurons and memristor-based spiking units [10, 11], pure SNNs typically encode information through binary spike events. This reduction in numerical precision complicates stable learning for robust dynamical-system tasks.

Recent progress has significantly improved SNN training by incorporating lessons from deep learning [12] and leveraging open-source machine learning infrastructures like SpikingJelly [13]. Innovations such as temporal dendritic heterogeneity [14], high-performance deep SNNs with extreme sparsity [15], and enhanced robustness paradigms [16] have substantially narrowed the performance gap. Furthermore, temporal processing modulated by neural oscillations [17], frugal SNNs for temporal pattern classification [18], memristive energy-efficient models [19], and advances in direct and rate-based backpropagation [20, 21] highlight the rapid maturation of spiking frameworks.

However, a persistent tension remains: ANNs preserve continuous-valued precision, whereas SNNs offer temporal sparsity but sacrifice graded signal representation. Complex-valued neural networks offer a relevant mathematical direction to resolve this, natively encoding information using magnitude and phase [22, 23]. These properties are particularly useful for dynamic applications such as event-based sensor fusion and mobile embodied perception [24, 25]. In parallel, SNNs are increasingly being applied to robust integrated estimation and cloud-edge event-driven control [26, 27], as well as advanced topologies including Riemannian manifolds and memristive graph learning [28, 29]. Yet, conventional complex-valued neurons do not inherently generate discrete, phase-gated events or impose biologically constrained spiking behaviors. Motivated by this gap, our recent biological study introduced the unified complex-valued neuron (UCN) as a phase-native single-neuron model [30]. In that formulation, the UCN explains how biological neurons jointly transmit "what" and "when" information through orthogonal magnitude and phase coordinates [30]. The magnitude represents graded signal strength, while the phase acts as a forced internal timing state whose temporal evolution inherently governs spike generation. In the present work, we extend the UCN from a biologically motivated single-neuron model into an engineered neural-network framework. We introduce the UCN-based Network (UCNN), where each neuron maintains an asymmetric complex-valued internal state. The magnitude channel provides bounded continuous activation, while the phase channel evolves according to driven temporal dynamics to gate event generation, effectively eliminating the need for complex, parallelized hybrid modules.

The main contributions of this paper are summarized as follows. First, we introduce the UCNN as a novel neural architecture; it is a magnitude-phase, event-valued computational model that provides continuous state accuracy while preserving sparse, phase-governed event timing. Second, we provide a unified training framework by formulating a foundational backpropagation and backpropagation through time (BP-BPTT) learning rule, which jointly optimizes the continuous magnitude pathway and the event-governed phase pathway end-to-end. Third, to reduce the computational burden of explicitly unrolling temporal dynamics, we introduce the event-driven adaptive phase learning (EAPL) rule. This ensures computational efficiency by propagating phase sensitivity through an adjoint-style recursion for highly efficient learning. Fourth, we provide spatiotemporal tracking validation by evaluating the UCNN on a mixed-modality object tracking task, demonstrating that the network achieves accurate spatiotemporal learning while preserving sparse event-driven computation. Finally, regarding dynamical system prediction, we validate the model's stability beyond visual perception by accurately predicting the chaotic, coupled nonlinear dynamics of the Lorenz attractor.

## 2. Theory

This section outlines the foundational concepts and mathematical notation necessary for the UCNN. It formalizes the structure of static and dynamic visual data, event-based representations, and standard neural processing paradigms prior to formally introducing the proposed neuron model.

***Static and dynamic visual sensing modalities***: Let $\mathrm{I_k} \in \mathrm{R}^{\mathrm{H\times W}}$ represent a standard camera frame at discrete time k, encoding absolute intensity to capture dense, low-frequency spatial structure and context. Conversely, event-based sensors generate discrete events only when local intensity changes exceed a threshold, isolating high-frequency temporal motion while filtering static redundancy. Consequently, robust perception in dynamic environments necessitates joint access to both complementary representations.

***Event-based representations from frame sequences***: In hybrid perception pipelines, event-like information is approximated from consecutive frames via a temporal difference map.

$$\mathrm{E_k} = |\mathrm{I_{k+1}} - \mathrm{I_k}| \quad (1)$$

Although synchronous at the sensor level, the resulting sparse matrix $\mathrm{E_k}$ retains essential event-driven characteristics (locality, sparsity, and temporal sensitivity) by highlighting pixel-level motion. Throughout this study, the frame sequence $\{\mathrm{I_k}\}$ supplies the static visual context, whereas the difference sequence $\{\mathrm{E_k}\}$ acts as the dynamic input stream encoding temporal structure.

***Continuous and event-driven neural processing paradigms***: Conventional ANNs process continuous inputs using layer-wise transformations of the form expresseion in Eq. (2).

$$\mathrm{u(t)} = \phi(\mathrm{W}\mathbf{x} + \mathrm{b}) \quad (2)$$

where $\mathbf{x}$ denotes the input vector (flattened data obtained from the image), W and b are trainable parameters, and $\phi(\cdot)$ is a nonlinear activation function. Although ideal for dense frames, evaluating every time step incurs high computational costs. Conversely, event-driven SNNs accumulate inputs and emit discrete outputs upon crossing thresholds, enabling sparse, energy-efficient processing but complicating gradient-based training and continuous-signal integration. HNNs attempt to bridge these paradigms using separate modules and hybrid units (HUs) [11]. However, this modular separation limits the tight coupling of static and dynamic streams and introduces significant design, training, and synchronization overheads.

***Unified representation of static and dynamic inputs***: To avoid redundant computational pathways in multimodal perception, dense static context and sparse dynamic updates must be jointly encoded. This requires an evolving internal state, $z_k$ at discrete time k, that accumulates both data types. Frame-based inputs drive the slow evolution of $z_k$ to maintain global context, whereas dynamic inputs induce rapid state transitions to capture instantaneous scene changes.

Formalized in the next section, this unified representation tightly couples static and dynamic information, enabling rapid event responsiveness alongside stable contextual awareness

***Problem setting and notation***: Let $\{t_k\}_{k=1}^{K}$ denote the discrete sampling instants of a frame-based camera. At each $t_k$, a dense image frame $I_k \in R^{H\times W}$ is acquired. No frame-based measurements are available during the inter-frame interval $(t_k, t_{k+1})$, thus in the other words the normal APS camera is blind there. Dynamic scene changes occurring within each inter-frame interval are acquired by DVS that by their nature sample the data with very higher frequencies and represented by a change map $E_k \in R^{H\times W}$ which summarizes intensity variations accumulated over $(t_k, t_{k+1})$. Note that, in this work, $E_k$ is being constructed from successive frames and serves as a proxy for event-driven sensing. The model operates on temporal windows of fixed length T. Within each window, inputs are presented sequentially as $\mathbf{x}_t \in R^{H\times W}$ where $\mathbf{x}_t$ corresponds either to a frame sample $I_k$ (provided by standard APS camera) or to an inter-frame dynamic update $E_k$ (provided form DVS), depending on the temporal index. Frame-based inputs appear sparsely in time, while dynamic updates dominate intermediate steps, reflecting the asynchronous sensing process. Given an input sequence $\{\mathbf{x}_t\}_{t=1}^{T}$, the learning objective is defined at the end of the window. In the implementations considered in this paper, the target output corresponds to the two-dimensional position of a moving object at the final frame time. The model must therefore maintain an internal state that evolves across inter-frame intervals and remains consistent with sparse frame observations. The neuron model introduced in the next section is designed to operate directly under this setting, supporting continuous internal state evolution across dynamic updates and periodic stabilization at frame times within a unified (neither parallelized nor modulated) computational computational framework.

### 3.1 The Unified Complex-valued Neural Network

This section mathematically formalizes the UCN at the neuron level, detailing its internal state representation, the functional roles of magnitude and phase, and the mechanism unifying continuous-valued and event-driven behaviors. Additionally, Figure 1 schematically compares the UCNN with traditional HNNs. The presented schematic in Figure 1(b) shows how the proposed UCNN in this study reduces the structural complexity and there is no need for synchronization, HUs, parallelization of two independent networks and their separate design and training rules, etc. consequently it will have improved efficiency while benefiting from semantic understanding of the data heritaged from ANNs.

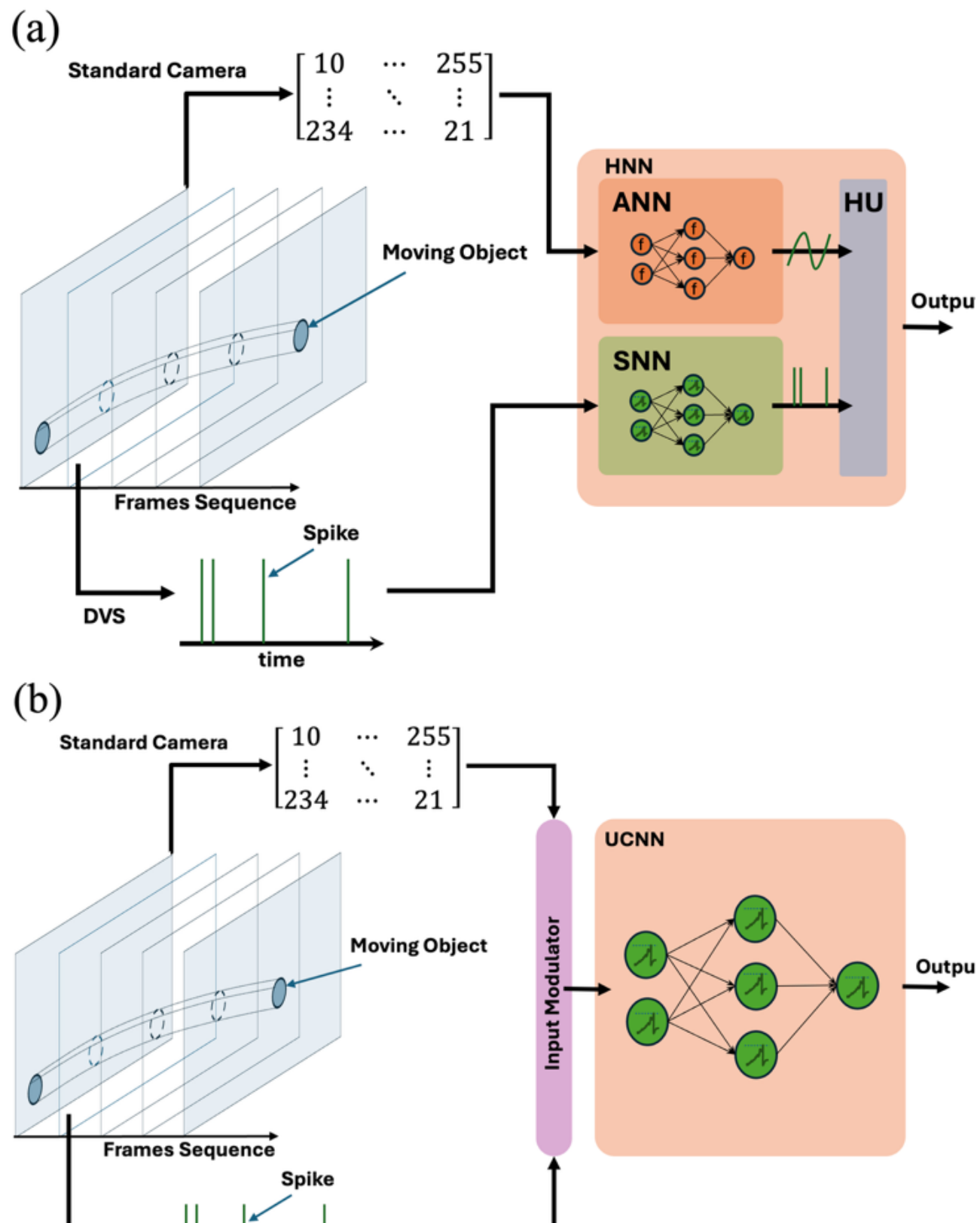

**Figure 1**. Schematic Demonstration of the UCNN Structure Compared with HNN, A) HNN, B) UCNN

***UCN State Representation***: The UCN is characterized by a complex-valued internal state defined Eq. (3) [30].

$$z(t) = r(t)e^{i\theta(t)} \quad (3)$$

where $r(t) \in R > 0$ denotes the magnitude component and $\theta(t) \in R$ denotes the phase component. Unlike conventional symmetric complex-valued models, the UCN assigns asymmetric roles to these components: magnitude encodes continuous signal strength, while phase acts as an internal timing variable governing discrete events. By gating the activation output (r) through the dynamical evolution of the phase (θ), this asymmetry allows a single internal state to simultaneously capture static contextual information and dynamic temporal structure.

***Pre-activation as a driving signal***: The pre-activation $u$ an affine transformation of the inputs, functions not as the final neuron output, but as a shared instantaneous stimulus driving both the magnitude and phase components. By coupling both channels to this single pre-activation, the UCN determines continuous activation and modulates internal timing through a unified representation, effectively integrating static and dynamic effects without requiring separate processing pathways.

***Magnitude channel (activation function)***: The magnitude channel produces a bounded continuous activation derived from the pre-activation $u(t)$. Specifically, the continuous activation is obtained using a hyperbolic tangent nonlinearity, and the effective magnitude used for event modulation is given by the rectified form expressed in Eq. (4).

$$r(t) = \max(0, \tanh(u(t))) \tag{4}$$

Ensuring biological plausibility, this formulation keeps the magnitude non-negative and bounded while retaining input sensitivity. Analogous to standard ANN activations, the magnitude channel encodes signal intensity to determine the specific value of each emitted event, seamlessly integrating with event-driven signaling.

***Phase channel dynamical representation***: The phase variable $\theta$ represents an internal dynamical state that evolves continuously over time. Its evolution follows a leaky, driven first-order dynamics analogous to that of a Leaky Integrate-and-Fire (LIF) neuron, expressed in Eq. (5).

$$\dot{\theta}(t) = -\lambda\theta(t) + \Omega_0 + u(t) \tag{5}$$

where $\lambda > 0$ is a leakage coefficient, $\Omega_0 > 0$ is a baseline angular frequency, and $u(t)$ is the time dependent pre-activation acting as an external drive. In this formulation, the leakage term prevents unbounded growth of the phase variable, while the baseline frequency ensures intrinsic progression of the internal timing state even in the absence of input. The pre-activation modulates the rate of phase evolution, allowing external signals to accelerate (excitation for positive pre-activation values) or decelerate (inhibition for negative pre-activation values) event generation. Crucially, the phase dynamics are independent of the magnitude activation, enabling the neuron to decouple event timing from signal amplitude. Further, since the event generation in the UCN is governed by the phase variable $\theta$ through a threshold-based gating mechanism. Here, $\theta_{th} = 2\pi$ denote the firing threshold [30]. A phase crossing event is detected when the internal phase exceeds this threshold. In its hard gating form, the spike indicator is defined as:

$$s(t) = H(\theta(t) - \theta_{th}) \tag{6}$$

Where $H(.)$ denotes the Heaviside step function. This formulation produces a binary event whenever the phase crosses the threshold and corresponds to strict event-driven behavior with no ambiguity in spike timing. To enable smoother transitions and facilitate gradient-based optimization, a soft gating formulation can alternatively be employed. In this case, the spike indicator is defined using a continuous surrogate function

$$s(t) = \sigma\left(\frac{\theta(t) - \theta_{th}}{\gamma}\right) \tag{7}$$

where $\sigma(\cdot)$ is a smooth nonlinear function, such as a clipped linear or sigmoid function, and $\gamma > 0$ controls the width of the transition region around the threshold. Soft gating allows the neuron to express partial activation near the firing boundary and provides non-zero gradients in the vicinity of threshold crossings.

***Phase reset mechanisms***: Following a threshold crossing, the phase variable must be updated to ensure continued cyclic evolution. Several reset strategies can be employed, each with distinct dynamical implications. In a hard reset-to-zero scheme, the phase is set explicitly to zero after a spike,

$$\theta(t^+) = 0 \tag{8}$$

where $t^+$ denotes the time immediately following event generation. This reset enforces a strict refractory cycle and results in highly regular phase trajectories. Alternatively, a modulo reset preserves excess phase accumulation by subtracting the threshold value,

$$\theta(t^+) = \theta(t) - \theta_{th} \tag{9}$$

This approach allows multiple threshold crossings within a short interval and supports higher firing rates under strong input drive, while maintaining continuity of the phase trajectory. On the other hand, in a soft reset formulation, the phase is smoothly attenuated rather than abruptly reset, for example by applying a decay factor,

$$\theta(t^+) = \rho\theta(t), \qquad 0 < \rho < 1 \tag{10}$$

Soft reset mechanisms reduce discontinuities in the phase dynamics and can improve numerical stability, particularly when combined with soft gating functions. The choice of reset mechanism determines how phase memory is preserved across events and directly influences firing regularity, responsiveness to strong inputs, and temporal smoothness of the neuron's internal dynamics.

***Valued event output***: The output of the UCN is defined by combining the phase-gated event indicator with the magnitude channel. The neuron output is given by Eq. (11).

$$y(t) = s(t) \cdot r(t) \tag{11}$$

Under hard gating, the neuron emits discrete valued events whose timing is determined solely by phase threshold crossings. Under soft gating, the output transitions smoothly near the threshold, allowing the neuron to interpolate between continuous and event-driven regimes. In both cases, the magnitude channel determines the value carried by each event, decoupling signal amplitude from event timing. Hard gating combined with reset-to-zero yields behavior most closely aligned with classical spiking neurons, producing sharply timed events and strictly separated firing cycles. Modulo reset enables higher temporal resolution by preserving excess phase and allowing rapid successive events. Soft gating and soft reset introduce smoothness into both event generation and internal state evolution, which can be advantageous for training and numerical robustness. Importantly, these mechanisms do not alter the fundamental structure of the UCN. Instead, they provide a flexible design space that allows the neuron to interpolate between fully discrete event-driven operation and hybrid continuous-event behavior, depending on application requirements and optimization constraints.

### 3.2 The Unified Learning Rule Development

This section derives a unified training rule within a single computational graph, jointly optimizing the network by combining conventional BP for the continuous magnitude pathway with BPTT and threshold surrogate gradients for temporal credit assignment in the event-governed phase pathway.

***Sequence model and readout***: Training spans temporal windows of length T, where at each internal step $t \in \{1, \dots, T\}$, the UCNN outputs a valued event (Eq. (11)) combining the phase-gated spike variable $s(t)$ and the rectified magnitude activation (Eq. (4)). In multilayer architectures, these events propagate forward as inputs to subsequent layers, with supervised regression utilizing the temporal mean of the final layer's aggregated output as the sequence-level readout expressed in Eq. (12).

$$\bar{r} = (1/T)\, \Sigma(t = 1 \text{ to } T)\, r(t) \tag{12}$$

followed by a linear output mapping

$$\hat{y} = W_{out} \left( \frac{1}{T} \sum_{t=1}^{T} s_2(t) r_2(t) \right) + b_{out} \tag{13}$$

This design makes the training objective depend on the entire temporal evolution of the UCNN state through the aggregated valued events.

***Composite loss and weighting***: Let y denote the target output and $\hat{y}$ denote the network prediction. A magnitude loss is defined by expression in Eq. (14).

$$\mathcal{L}_{mag} = \left(\frac{1}{2}\right) \|\hat{y} - y\|_2^2 \tag{14}$$

To explicitly control the relative influence of value-driven versus timing-driven gradients, a weighting parameter $\beta_{val} > 0$ scales the magnitude loss, yielding

$$\mathcal{L} = \beta_{val}\, \mathcal{L}_{mag} \tag{15}$$

Timing sensitivity is introduced through the gradient pathways that pass-through spike gating and phase evolution, scaled by a separate coefficient $\alpha_{time} > 0$, as detailed below. This separation is consistent with the "what vs. when" optimization objective described in the problem assumptions.

***Surrogate gradient for phase-gated spikes***: Under hard gating (Eq. (6)), the derivative $\partial s/\partial \theta$ is zero almost everywhere. During training, a surrogate function is introduced near threshold crossings. A triangular surrogate consistent with the implementation is expressed in Eq. (16).

$$\frac{\partial s}{\partial \theta} \approx \psi_\gamma(\theta - \theta_{th}) = \left(\frac{1}{\gamma}\right) \max\left(0, 1 - \frac{|\theta - \theta_{th}|}{\gamma}\right) \tag{16}$$

The above expression is applied elementwise to the phase vector, where $\gamma > 0$ controls the width of the nonzero-gradient region. Using this surrogate, the timing-driven contribution to the gradient of phase at step t becomes

$$\partial\mathcal{L}/\partial\theta(t)|_{spk} = \alpha_{time}\, (\partial\mathcal{L}/\partial s(t) \odot \psi_\gamma(\theta(t) - \theta_{th})) \tag{17}$$

This term assigns credit to phase trajectories that move the gate toward or away from threshold in a way that reduces the task loss.

***BPTT through phase evolution and coupling to pre-activation***: The phase dynamics were defined earlier as a leaky driven process. For learning, what matters is that $\theta(t + 1)$ depends on $\theta(t)$ and on the pre-activation $u(t)$. Denoting the phase update map abstractly by Eq. (18).

$$\theta(t + 1) = F(\theta(t), u(t)) \tag{18}$$

BPTT accumulates gradients backward through time via

$$\frac{\partial\mathcal{L}}{\partial\theta(t)} = \frac{\partial\mathcal{L}}{\partial\theta(t)|_{spk}} + (\partial F/\partial\theta(t))^T\, \partial\mathcal{L}/\partial\theta(t + 1) \tag{19}$$

The timing gradient then induces an additional contribution to the pre-activation gradient through the phase dependence on $u(t)$:

$$\partial\mathcal{L}/\partial u(t)|_{time} = (\partial F/\partial u(t))^T\, \partial\mathcal{L}/\partial\theta(t + 1) \tag{20}$$

In the implemented discrete-step realization, this coupling is approximated by a constant scale proportional to the internal step size, yielding the practical form $\partial\mathcal{L}/\partial u(t)|_{time} \propto \partial\mathcal{L}/\partial\theta(t + 1)$. This approximation is precisely what makes the phase pathway trainable without requiring explicit differentiation through a particular numerical solver, while still assigning temporal credit through the unrolled phase trajectory.

***Unified parameter gradients for a UCN layer***: Combining both pathways yields the total pre-activation gradient at each time step:

$$\partial \mathcal{L}/\partial \mathrm{u(t)} = \partial \mathcal{L}/\partial \mathrm{u(t)}|_{\mathrm{val}} + \partial \mathcal{L}/\partial \mathrm{u(t)}|_{\mathrm{time}} \quad (21)$$

For a layer with affine pre-activation $\mathrm{u(t)} = \mathrm{W}\mathbf{x}(\mathrm{t}) + \mathrm{b}$, the parameter gradients follow as

$$\frac{\partial \mathcal{L}}{\partial \mathrm{W}} = \frac{\Sigma(\mathrm{t} = 1 \text{ to } \mathrm{T})\, \partial \mathcal{L}}{\partial \mathrm{u(t)x(t)^T}} \quad (22)$$

$$\partial \mathcal{L}/\partial \mathrm{b} = \Sigma(\mathrm{t} = 1 \text{ to } \mathrm{T})\ \partial \mathcal{L}/\partial \mathrm{u(t)} \quad (23)$$

This expression makes explicit that the UCN learning rule is unified at the neuron level: a single synaptic weight contributes simultaneously to the continuous magnitude computation and to the event-timing computation through the phase pathway.

***Extension to multilayer UCNN***: For multilayer architectures, the above derivation applies to each layer, with the difference that the "input" $x(t)$ to a deeper layer is itself a valued event vector produced by the preceding layer. The backward pass therefore propagates both value-driven gradients (through the tanh magnitude path) and timing-driven gradients (through the phase gate and unrolled phase state) layer by layer, producing unified updates for $W\ell, b\ell$ at each layer $\ell$.

***Inference-time hard gating versus training-time soft gradients***: During inference, the spike gate is typically implemented as hard thresholding to preserve event sparsity and deterministic firing. During training, the surrogate derivative is used only to define $\partial s/\partial \theta$ in the backward pass; the forward pass may still operate with hard gating. This separation preserves the desired event-driven behavior while enabling end-to-end optimization through gradient-based methods. Although the BP-BPTT formulation provides a unified optimization framework for both magnitude and phase pathways, it requires explicit temporal unrolling of the phase dynamics and separate treatment of value and timing gradient contributions. Consequently, computational cost increases linearly with sequence length and network depth. To address this limitation, a novel alternative event-driven adjoint formulation is introduced in the next section.

### 3.3 Event-driven adjoint phase-learning

Although BP-BPTT unifies training, explicitly unrolling phase dynamics and separately propagating gradients incurs high computational costs as networks scale. To mitigate this, we introduce event-driven adjoint phase learning (EAPL), which circumvents explicitly unrolled computational graphs by reformulating learning as a backward dynamical process. By utilizing an auxiliary adjoint state that evolves backward in time to accumulate task-relevant error, EAPL significantly reduces the computational burden of conventional BP-BPTT while remaining fully compatible with the UCNN architecture.

***Phase dynamics and adjoint state***: EAPL introduces an adjoint state expressed in Eq. (24).

$$a(t) = \partial L / \partial \theta(t) \quad (24)$$

which measures the sensitivity of the loss to the phase trajectory. For a general phase system $\mathrm{d\theta/dt = f(\theta, u)}$, the continuous adjoint evolves as

$$\mathrm{da(t)/dt} = -(\partial \mathrm{f}/\partial \theta)^{\mathrm{T}}\, \mathrm{a(t)} \quad (25)$$

For the UCN phase equation, $\partial \mathrm{f}/\partial \theta = -\lambda$, so the backward phase sensitivity follows the simple linear rule

$$\mathrm{da(t)/dt} = \lambda \mathrm{a(t)} \quad (26)$$

Between spike events, the backward error signal propagates directly through the internal phase dynamics, bypassing the explicit unfolding of intermediate operations required by conventional BPTT. Spike generation remains governed by the phase threshold, with the non-differentiable derivative of the Heaviside function replaced by a local surrogate function around the threshold boundary during training:

$$\psi(\theta) = \max(0, 1 - |\theta - \theta_{\mathrm{th}}|/\gamma)/\gamma \quad (27)$$

At each event or near-threshold crossing, the adjoint receives a timing correction expressedn below:

$$\mathrm{a} \leftarrow \mathrm{a} + \alpha_{\mathrm{time}}\, (\partial \mathrm{L}/\partial \mathrm{s})\, \psi(\theta) \quad (28)$$

This term injects temporal credit into the adjoint variable only where spike timing is relevant. The magnitude channel remains governed by $\mathrm{q(t) = max(0, tanh(u(t)))}$, and the valued event is $r(t) = s(t)q(t)$. The total pre-activation gradient combines the magnitude contribution and the phase-adjoint contribution. For a discrete implementation over a time window of length T, the update can be written compactly as

$$\delta_{\mathrm{t}} = \delta_{\mathrm{t}}^{\mathrm{value}} + \delta_{\mathrm{t}}^{\mathrm{phase}} \quad (29)$$

$$\frac{\partial \mathrm{L}}{\partial \mathrm{W}} = \Sigma_{\{t=1\}}^{\{T\}}\, \delta_{\mathrm{t}}\, \mathrm{x_t}^{\mathrm{T}} \quad (30)$$

$$\frac{\partial \mathrm{L}}{\partial \mathrm{b}} = \Sigma_{\{t=1\}}^{\{T\}}\, \delta_{\mathrm{t}} \quad (31)$$

In the UCNN used in this paper, the same rule is applied first to the second layer using $r_1(t)$ as input, and then to the first layer using $x(t)$ as input. The output layer is trained by the standard regression gradient from $\hat{y} = W_{out}\, mean(r^2) + b_{out}$. From a computational perspective, EAPL transforms UCNN training into a backward-time dynamical process. The forward phase trajectory determines when each neuron emits valued events, while the backward adjoint trajectory propagates task-relevant sensitivity information through the same phase dynamics. Event-driven corrections occur near threshold crossings, allowing spike timing to influence learning without requiring a separate timing-gradient sweep. As a result, EAPL preserves the continuous-valued magnitude pathway, the event-driven phase pathway, and the sparse firing behavior of UCNNs while reducing the redundant backward computations used in BP-BPTT

***Discrete EAPL recursion UCNN dynamics:*** For the implemented UCN phase equation, the sensitivity of the pre-reset phase with respect to the previous phase and pre-activation can be approximated using the exact solution of the linear phase equation over one internal step:

$$A = \exp(-\lambda\Delta t), \tag{32}$$

$$B = (1 - \exp(-\lambda\Delta t))/\lambda,\ \ \lambda > 0, \tag{33}$$

with $B = \Delta t$ when $\lambda = 0$. These coefficients represent

$$\frac{\partial\theta_{pre,t+1}}{\partial\theta_t} \approx A \tag{34}$$

$$\frac{\partial\theta_{pre,t+1}}{\partial u_t} \approx B \tag{35}$$

The backward phase adjoint recursion is then written as

$$a_t = A\,\mu_t \tag{36}$$

and the timing contribution to the pre-activation gradient becomes

$$\left(\frac{\partial L}{\partial u_t}\right)_{time} = B\,\mu_t \tag{37}$$

The value contribution is obtained from the magnitude channel. with $v_t = \tanh(u_t)$, $q_t = \max(0, v_t)$, $r_t = s_t \odot q_t$ the value-path pre-activation gradient is:

$$\left(\frac{\partial L}{\partial u_t}\right)_{value} = [(\partial L/\partial r_t) \odot s_t] \odot I(v_t > 0) \odot (1 - v_t^2) \tag{38}$$

The total EAPL pre-activation gradient is therefore:

$$\delta_t = \left(\frac{\partial L}{\partial u_t}\right)_{value} + \left(\frac{\partial L}{\partial u_t}\right)_{time} \tag{39}$$

For a layer with affine pre-activation $u_t = Wx_t + b$, the parameter updates are

$$\frac{\partial L}{\partial W} = \Sigma_{\{t=1\}}^{\{T\}}\, \delta_t\, x_t^{\,T} \tag{40}$$

$$\frac{\partial L}{\partial b} = \Sigma_{\{t=1\}}^{\{T\}}\, \delta_t \tag{41}$$

For deeper UCNN layers, $x_t$ is replaced by the valued event vector from the previous layer. This preserves the same learning structure across layers.

3.4 Computational complexity analysis

The proposed EAPL learning rule preserves the same asymptotic training complexity class as the previous BP-BPTT based implementation, namely $O(T(H_1D + H_1H_2))$, where T is the sequence length, D is the input dimension, and $H_1, H_2$ are the hidden-layer sizes. However, EAPL reduces the dominant backward-pass computational cost from $O(T(2H_1D + 4H_1H_2))$ to $O(T(H_1D + 2H_1H_2))$ by replacing separate value and timing backward sweeps with a unified adjoint-style recursion. For example in the object tracking problem for the network used in this study the values of $D = 1024, H_1 = 128, H_2 = 64$, and $T = 11$ which corresponds to an approximate 2 times reduction in dominant backward operations, while retaining the same asymptotic memory complexity $O(T(H_1 + H_2))$.

## 4 Numerical Simulations

Before presenting the task-driven assessment of the UCNN, we first investigate the intrinsic dynamical behavior of the proposed network. The objective of theb first analysis is not performance evaluation, but rather to qualitatively demonstrate how the proposed neuron model responds to mixed static and dynamic inputs, how activity propagates across layers, and how phase-driven event generation emerges naturally from the input structure.

4.1 Case Study 1: network behavior analysis

Here, at this step, a two-layer UCNN is constructed with no learning and no parameter adaptation. The first layer consists of 1024 neurons, corresponding to the flattened spatial dimension of the input frames (32×32 images), while the second layer contains 512 neurons receiving weighted valued-event outputs from the first layer through a randomly initialized dense weight matrix. All synaptic weights are drawn from a zero-mean normal Gaussian distribution $N(0,1)$ and remain fixed throughout the simulation. Note that in the UCNN the positive synaptic weights represent the excitatory synapses, and the negative weights represent the inhibitory synapses. Each neuron is initialized with zero phase and zero magnitude. Neuron-specific intrinsic angular frequencies are randomly assigned for both layers that are absolute values of random samples from a normal Gaussian distribution $N(0,1)$. The network is driven by the same mixed-modality input stream used throughout the paper, consisting of sparse static frames and inter-frame dynamic change maps. Specifically, a static frame is injected periodically, while intermediate time steps are driven by dynamic error maps that encode motion-related changes. This input scheduling ensures that the network alternates between context-setting inputs and high-temporal-resolution dynamic stimulations. The internal phase dynamics of each neuron evolve continuously according to the UCN phase model expressed in Eq. (5), while spike events are generated when the phase completes a full $2\pi$ rotation. Upon threshold crossing, the neuron emits a valued event whose magnitude is given by a rectified hyperbolic tangent of the instantaneous input expressed in Eq. (4). Sub-figures in Figure 2 corresponding to the first and second UCNN layers show the spatiotemporal distribution of valued spike events over the simulation horizon. Each horizontal index corresponds to an individual neuron, while the horizontal axis represents discrete internal time steps. In the first layer, spiking activity exhibits a clear structured pattern that reflects the spatiotemporal content of the input stream. Periods dominated by dynamic inputs produce sparse, localized spiking, while the injection of static frames leads to synchronized vertical activation bands across large neuron subsets. This behavior emerges despite the absence of learning and indicates that the phase dynamics naturally synchronize in response to global input changes while remaining sparse during inter-frame motion-driven stimulation. In contrast, the second layer exhibits more distributed and decorrelated activity. The random projection of first-layer valued events through the fixed weight matrix spreads activation across neurons and time, producing a richer temporal pattern with reduced synchrony. This transformation illustrates how deeper UCNN layers can progressively reshape spatiotemporal activity while preserving event-driven sparsity.

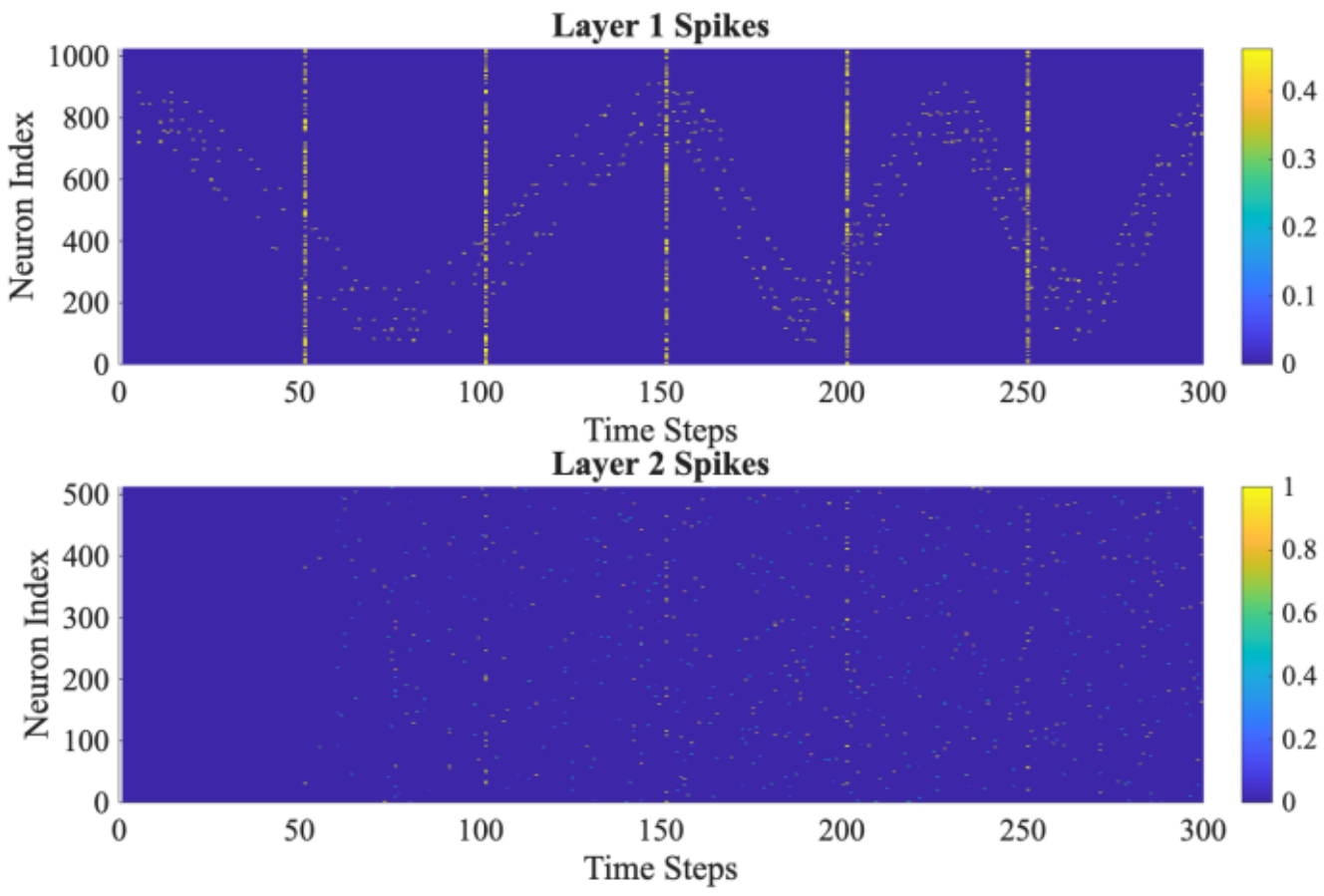

**Figure 2**. Obtained spiking patterns for two layers of the UCNN network

Quantitatively, the total number of spike events remains a small fraction of all possible neuron-time combinations, confirming that the network operates in a sparse, event-driven regime even under continuous stimulation. This sparsity is achieved without any explicit regularization or threshold tuning, arising purely from the phase-gated neuron model.

To further illustrate the internal dynamics of individual neurons, the phase trajectory of a randomly selected neuron from the second layer is illustrated in Figure 3. The phase evolves smoothly over time, exhibiting approximately linear growth modulated by the input drive. When the phase reaches the $2\pi$ threshold, a spike event is generated and the phase is reset modulo the threshold, producing the characteristic sawtooth-like trajectory.

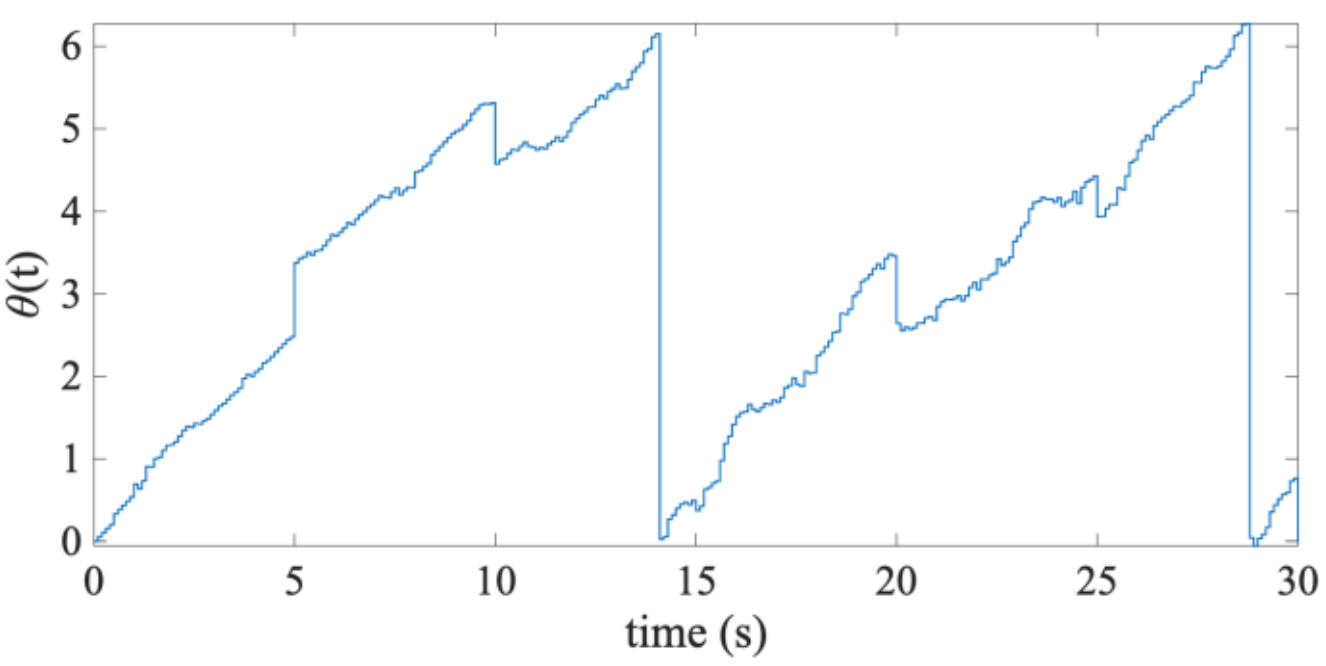

**Figure 3**. Obtained phase variation for a randomly selected UCN neuron in the layer two.

These phase variations provide direct evidence that spike generation in the UCNN is governed by continuous internal dynamics rather than instantaneous thresholding of the input. Importantly, variations in the slope of the phase evolution correspond to changes in the input signal, demonstrating how dynamic stimuli modulate event timing without altering the magnitude computation pathway. This implementation demonstrates that the UCNN exhibits meaningful and interpretable spatiotemporal behavior even in the absence of learning. The combination of continuous phase evolution, threshold-based event generation, and valued spike outputs leads to structured activity that reflects both static context and dynamic motion in the input stream. The clear separation between synchronized frame-driven responses and sparse motion-driven activity highlights the suitability of the UCNN for mixed-modality perception tasks.

These results establish a mechanistic foundation for the subsequent learning-based utilizations. In the following sections, the same neuron model and input structure are used to perform supervised object tracking, enabling direct comparison between ANN, SNN, and UCNN architectures using the identical problem with their own consistent dataset.

### *4.2 Case study 2: Object tracking problem*

The In this case study, the UCNN is evaluated on an object tracking task using the moving object on circular path dataset with frames of size 32×32 pixels. The objective is to assess whether the proposed neuron model, when trained end-to-end using the unified learning rule, can accurately estimate the two-dimensional position of a moving object from a mixed stream of static and dynamic visual inputs. Here, the dataset consists of grayscale image frames depicting a circular object moving along a smooth $T = 11$. For each window, the input sequence is formed by selectively injecting either a static frame or a dynamic error map at each internal time step. The first frame in each window initializes the context, while subsequent steps are predominantly driven by dynamic error maps, with periodic reinsertion of static frames to refresh global spatial information. This scheduling directly reflects the asynchronous sensing assumption, where dense frame information is sparse in time and fast dynamics are captured through inter-frame updates. The target associated with each input window is the ground-truth object center $(x, y)$ corresponding to the final frame of the window. This formulation requires the network to integrate information over time and maintain a coherent internal state across both static and dynamic inputs. All inputs are normalized globally using the mean and standard deviation computed over the full training set, ensuring stable optimization

Figure 4 summarizes the comparative performance of the ANN, SNN, and proposed UCNN architectures on the same moving-object tracking scenario. The objective of this comparison is to evaluate the three models under a controlled setting using the same target trajectory and comparable network dimensions, while highlighting the different computational principles underlying each architecture. The ANN represents the continuous-valued frame-based baseline, the SNN represents the purely event-driven baseline, and the UCNN represents the proposed magnitude-phase event-valued model. Figure 4(a) shows the training loss histories in terms of average MSE loss. All three models exhibit a rapid reduction in loss during the initial epochs, followed by a slower convergence phase. The ANN converges to the lowest final loss, which is expected because it receives dense frame-based spatial information and directly learns the mapping from image intensity to object position. The SNN also reduces its training loss but converges to a higher error level and displays a less smooth loss profile, reflecting the difficulty of optimizing binary spike-based dynamics through BPTT and surrogate gradients. The UCNN converges stably throughout training and achieves a final loss between the ANN and SNN. This behavior indicates that the proposed UCNN learning framework successfully propagates error information through both the continuous magnitude pathway and the phase-driven event pathway, while avoiding the stronger instability observed in the purely spiking baseline. Figure 4(b) provides a component-wise RMSE comparison for the x-coordinate, y-coordinate, and overall two-dimensional position error. The ANN produces the smallest RMSE values in all components, confirming its strong performance when full frame-based information is available at each prediction step. The SNN exhibits the largest RMSE values, indicating that temporal difference information alone is insufficient to recover the object center with the same precision as dense spatial input. In contrast, the UCNN considerably reduces the error relative to the SNN in both coordinate directions and in the two-dimensional RMSE. This confirms that combining continuous-valued signal representation with phase-gated event timing improves tracking accuracy compared with a purely event-driven SNN. Figure 4(c) presents the cumulative distribution function of the tracking error. The ANN curve rises sharply near zero, showing that most ANN predictions have very small tracking error. The SNN curve is shifted to the right and has a slower rise, indicating a broader error distribution and more frequent large deviations. The UCNN curve lies between the ANN and SNN but is substantially closer to the ANN, demonstrating that most UCNN predictions remain within a relatively small error range. This distribution-level comparison is important because it shows not only the average performance, but also the consistency of each model across the full test sequence. Figure 4(d) reports the mean tracking error together with standard deviation and maximum observed error. The ANN has the smallest mean error and the narrowest variability, consistent with its dense-frame advantage. The SNN has the largest mean error, the largest standard deviation, and the highest maximum error, confirming that its predictions are more sensitive to temporal ambiguity and spike-based optimization fluctuations. The UCNN significantly lowers both the mean error and variability compared with the SNN, while also reducing the maximum tracking error. These results demonstrate that the UCNN provides a more stable tracking behavior than the SNN and avoids excessive jitter by allowing each event to carry a continuous-valued magnitude in addition to phase-driven timing.

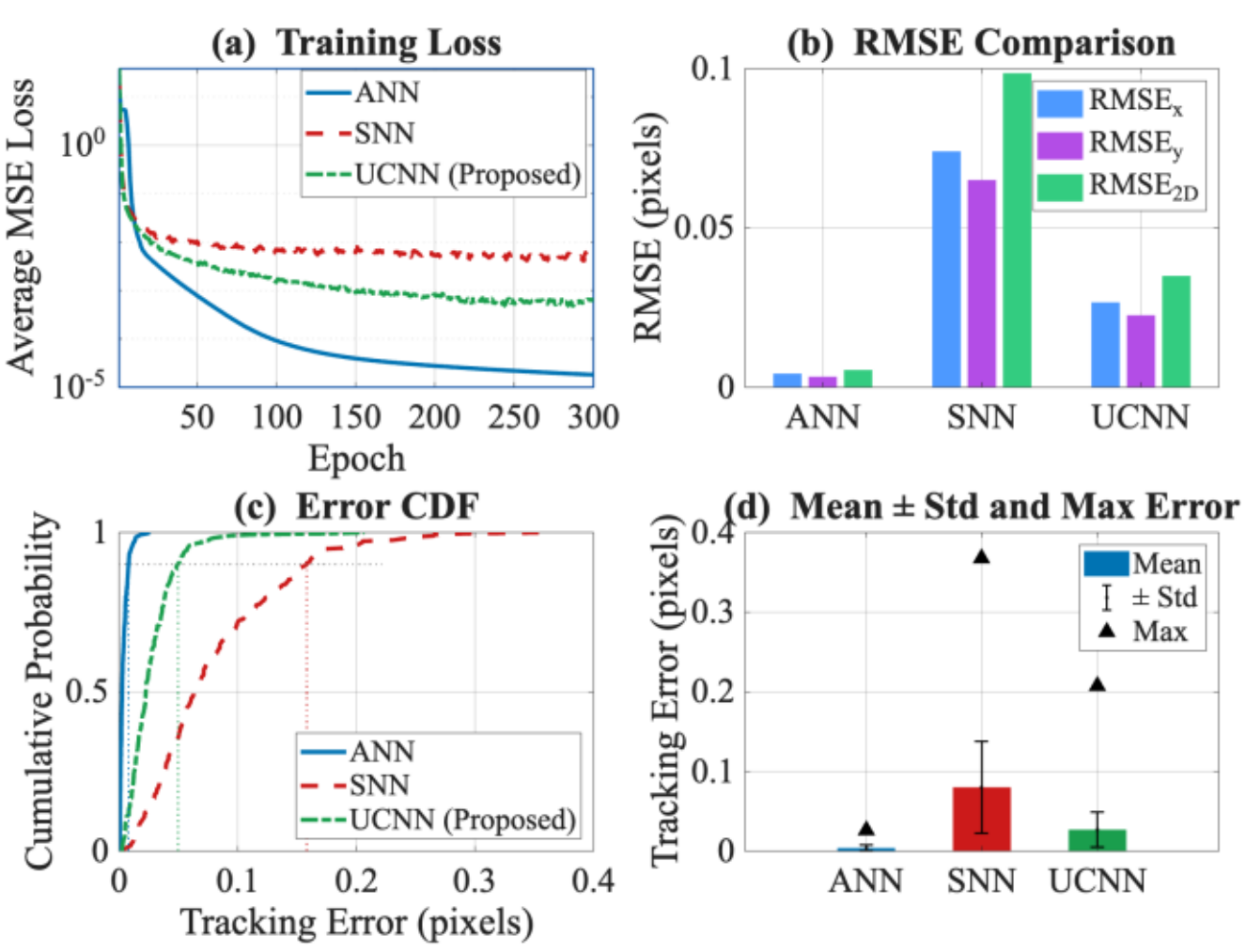

**Figure 4**. Comparative performance evaluation of ANN, SNN, and the proposed UCNN on the moving-object tracking task. (a) Training loss histories for the three models in terms of average MSE loss. (b) Root-mean-square error comparison along the x-direction, y-direction, and two-dimensional position error. (c) Cumulative distribution function of tracking error for the three architectures. (d) Mean tracking error with standard deviation and maximum observed error. The ANN achieves the lowest error due to direct access to dense frame-based spatial information, while the SNN exhibits the largest error and variability because it relies only on event-driven temporal information. The proposed UCNN substantially reduces the SNN tracking error and provides a more stable event-driven representation by jointly encoding continuous-valued magnitude and phase-driven temporal dynamics.

Overall, the comparison in Figure 4 demonstrates the expected trade-off among the three neural computation paradigms. The ANN provides the highest numerical accuracy under dense frame availability, but it does not offer intrinsic event-driven temporal computation. The SNN provides sparse event-driven processing, but its binary spike representation leads to lower tracking precision and higher variability. The proposed UCNN occupies an intermediate and practically important regime: it preserves event-driven temporal dynamics while carrying continuous-valued information through its magnitude channel. Therefore, the UCNN substantially improves upon the SNN baseline and provides a more stable and expressive event-driven architecture for spatiotemporal tracking. Figure 5 compares the ground-truth trajectory of the moving object with the trajectories predicted by the ANN, SNN, and proposed UCNN models. The purpose of this figure is to provide a qualitative assessment of the tracking behavior of the three architectures over the complete motion cycle. As shown, all three models successfully reconstruct the overall circular trajectory, indicating that each architecture is capable of learning the dominant motion pattern in the considered tracking scenario. The ANN-predicted trajectory nearly overlaps with the ground-truth curve throughout the full cycle. This behavior is consistent with the quantitative results reported earlier, where the ANN achieved the lowest tracking error due to its direct access to dense frame-based spatial information. The SNN also captures the global trajectory shape; however, small local deviations can be observed along the path.

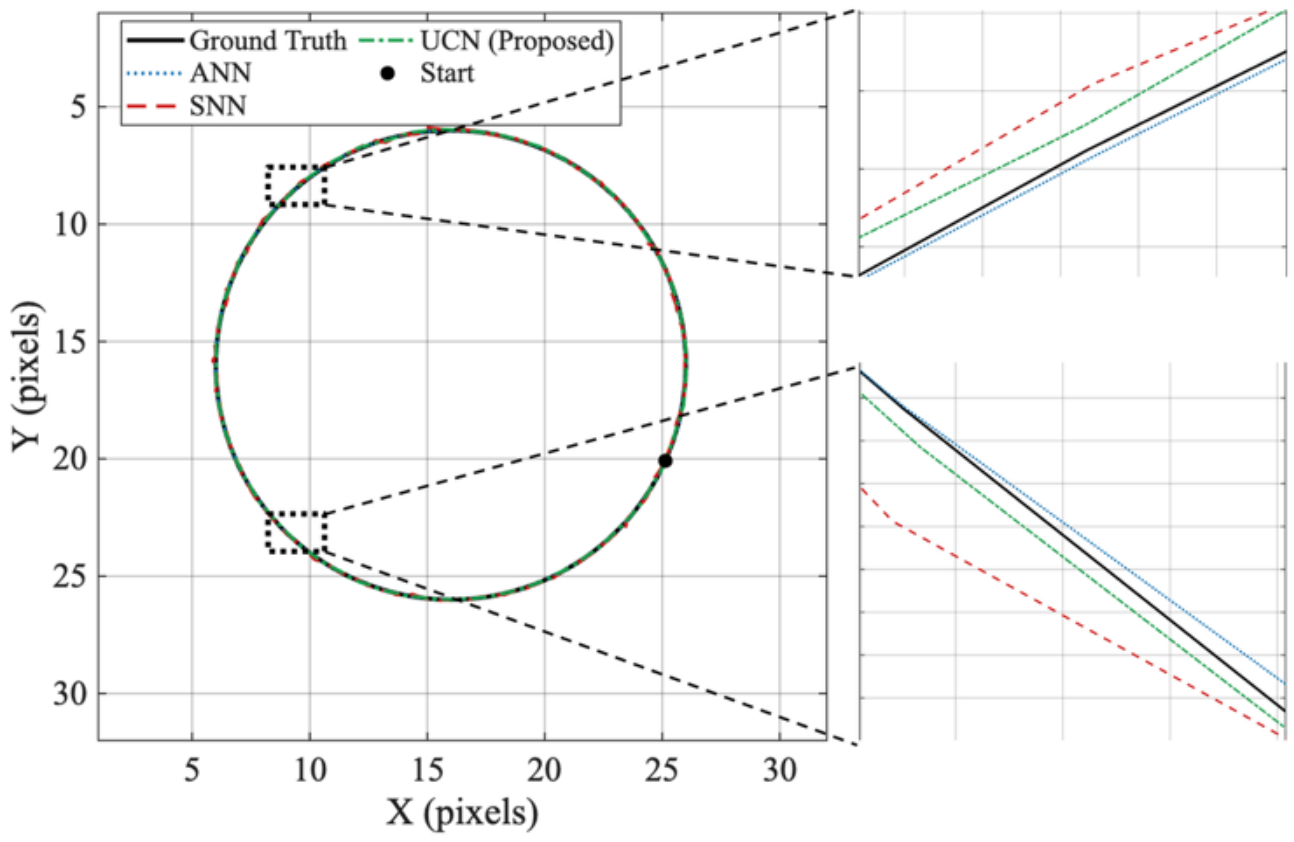

**Figure 5**. Comparison of ground-truth and predicted object trajectories for the ANN, SNN, and proposed UCNN models. The black solid curve represents the ground-truth circular trajectory, while the blue dotted, red dashed, and green dash-dotted curves correspond to the ANN, SNN, and UCNN predictions, respectively. The black marker indicates the starting point of the trajectory.

These deviations are consistent with the higher RMSE, broader error distribution, and larger maximum error obtained for the SNN, and reflect the limitations of relying only on event-driven temporal information without continuous-valued spatial context. The proposed UCNN trajectory closely follows the ground truth and remains visibly more consistent than the SNN prediction. This result demonstrates that the UCNN can preserve the temporal advantages of event-driven processing while improving tracking stability through its continuous magnitude channel. In other words, the phase component supports event timing and temporal evolution, while the magnitude component carries graded signal information that helps reduce the ambiguity associated with purely binary spike-based representations. Therefore, the trajectory-level comparison confirms that the UCNN provides a more stable and accurate event-driven tracking behavior than the SNN baseline, while approaching the dense-frame ANN trajectory. Figure 6 presents the smoothed tracking-error histories obtained from the ANN, SNN, and proposed UCNN models over the complete test trajectory. This figure complements the trajectory plot by showing how the prediction error evolves with respect to the frame index, rather than only comparing the geometric overlap between the predicted and true paths. The ANN maintains the lowest error throughout the sequence, with the smoothed error remaining close to zero for almost all frames. This result is consistent with the previous observations that the ANN provides the strongest numerical accuracy when dense frame-based inputs are continuously available. The small variations in the ANN curve indicate minor frame-to-frame sensitivity caused by image discretization and local changes in the object silhouette, but these fluctuations remain limited. The SNN exhibits the largest error over the entire sequence. Its error curve shows stronger temporal fluctuations and repeated peaks, indicating that the purely event-driven representation is more sensitive to ambiguity in the temporal difference maps. Since the SNN relies primarily on spike timing and rate information without direct access to dense spatial context, it has reduced ability to recover the absolute object position when the dynamic information is weak or locally ambiguous.

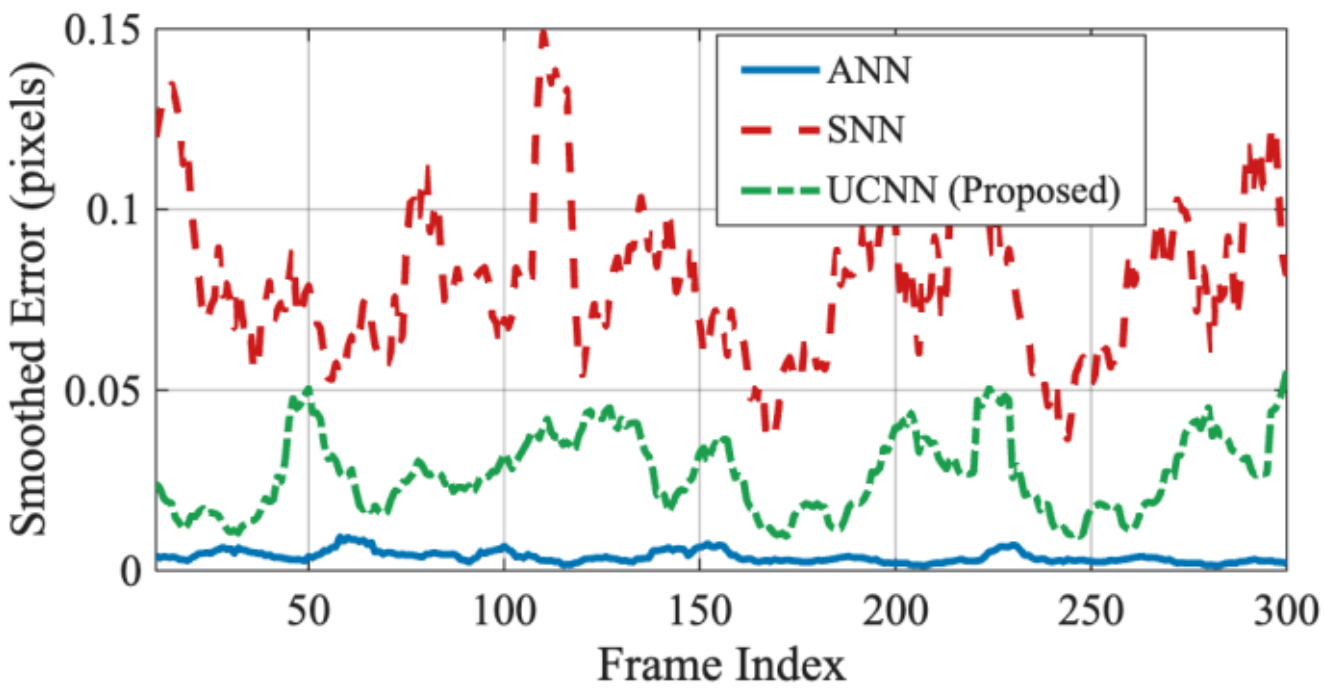

**Figure 6**. Smoothed tracking-error histories of the ANN, SNN, and proposed UCNN models over the test trajectory. The curves report the frame-wise Euclidean tracking error after smoothing to emphasize the overall error trend and temporal stability of each architecture.

This behavior demonstrated in Figure 6 explains the higher RMSE and broader error distribution reported in the quantitative comparison. The proposed UCNN produces an error profile between the ANN and SNN, while remaining substantially closer to the ANN than to the SNN. The UCNN curve shows that the tracking error remains bounded and significantly lower than that of the SNN across the full sequence. This improvement demonstrates that the

magnitude–phase formulation provides a more stable event-driven representation: the phase channel preserves temporal/event-based dynamics, while the magnitude channel carries continuous-valued information that improves positional accuracy. Therefore, the smoothed error history confirms that the UCNN reduces the tracking jitter and error variability associated with the purely spiking baseline while retaining event-driven temporal processing.

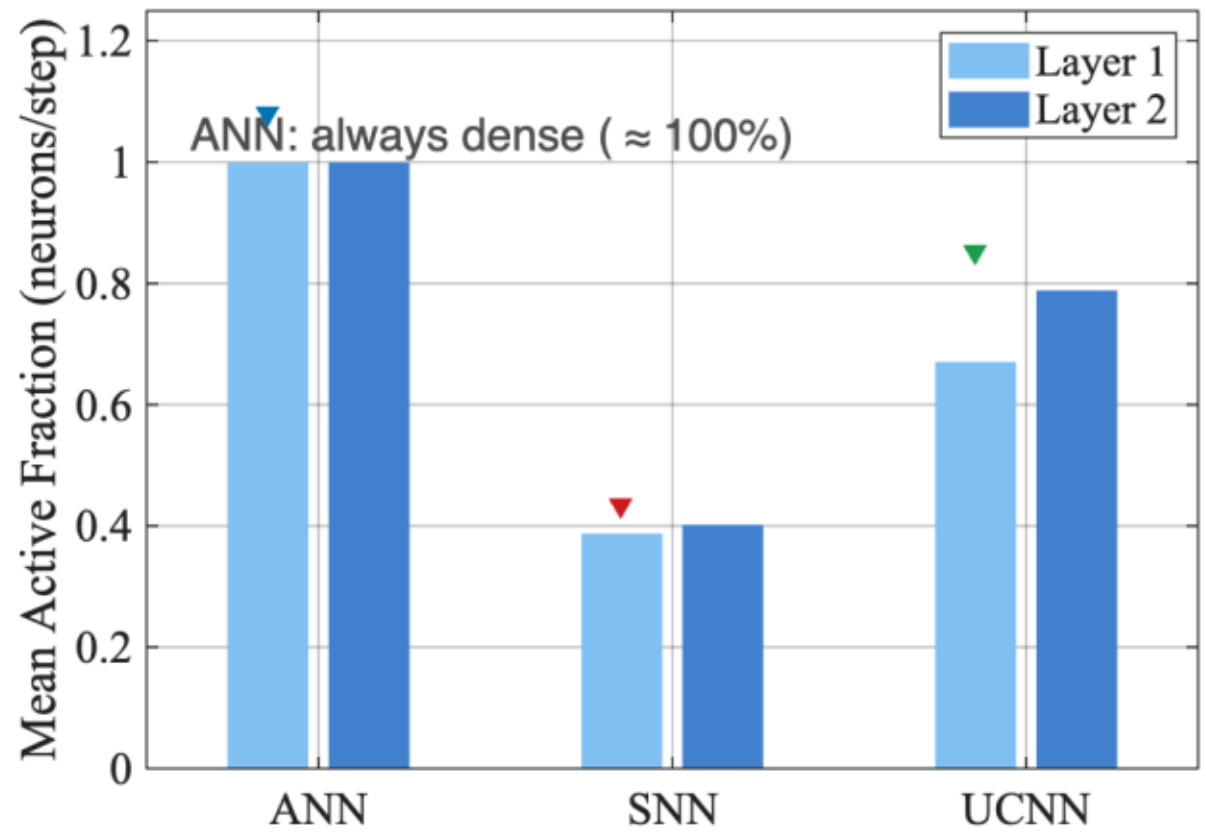

**Figure 7**. Mean active fraction of neurons per step for ANN, SNN, and UCNN across two hidden layers.

Figure 7 compares the average neural activity of the ANN, SNN, and proposed UCNN models. The ANN remains fully dense in both layers, meaning that almost all neurons are active at every inference step. In contrast, the SNN exhibits the lowest activity because computation is triggered only by spike events. The proposed UCNN shows an intermediate activity level: it is substantially sparser than the ANN while remaining more active than the purely spiking SNN. This behavior reflects the UCNN's unified computational nature, where phase-gated event generation preserves sparsity while magnitude-valued events retain richer information flow. Therefore, the UCNN provides a practical balance between dense ANN-style representation and sparse SNN-style event-driven computation. Figure 8 compares the post-training event activity of the SNN and the proposed UCNN in the first and second hidden layers. The SNN exhibits irregular spike patterns, especially in the first layer, where activity is distributed in a highly scattered manner across neurons and time. In the second SNN layer, the activity becomes more structured, but it still reflects binary spike-based communication driven only by threshold crossings. In contrast, the UCNN produces more organized and temporally consistent event patterns in both layers. This behavior is a direct consequence of the UCNN magnitude-phase formulation, where the phase governs event timing while the magnitude assigns a continuous value to each emitted event. Therefore, the UCNN does not rely only on binary spike occurrence, but preserves richer information flow through valued events. The structured activity observed in both UCNN layers indicates that training shapes the phase dynamics into stable task-relevant temporal patterns, supporting more accurate and less jitter-prone tracking compared with the purely spiking SNN baseline.

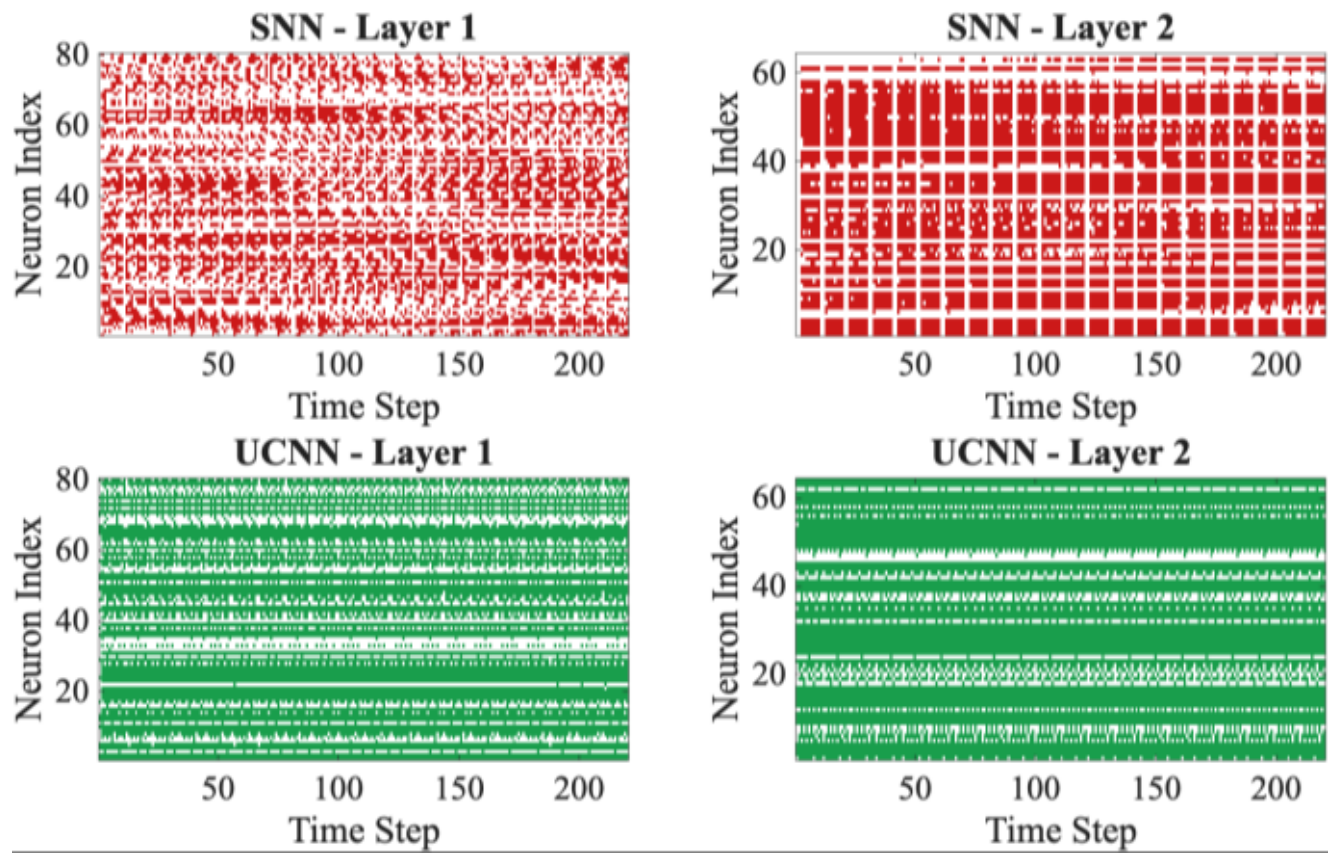

**Figure 8**. Post-training spike activity comparison between the SNN and the proposed UCNN across two hidden layers. The upper panels show the spike rasters of the SNN, while the lower panels show the corresponding event activity of the UCNN after training

*4.3 Case study 3: Lorenz attractor*

In this case study, the proposed UCNN is evaluated on the well-known Lorenz attractor as a nonlinear chaotic dynamical-system learning problem. The objective is to assess whether the proposed magnitude-phase neuron model can learn coupled temporal dynamics beyond the visual object-tracking scenario. The Lorenz system is defined as

$$\begin{aligned} \dot{x} &= \sigma(y - x) \\ \dot{y} &= x(\rho - z) - y \\ \dot{z} &= xy - \beta z \end{aligned} \tag{42}$$

where $x, y, and\ z$ are the system states, $\sigma$ is the Prandtl number, $\rho$ is the Rayleigh parameter, and $\beta$ is the geometric factor. In this study, the standard chaotic parameters are used: $\sigma = 10$, $\rho = 28$, and $\beta = 8/3$. The generated Lorenz trajectory is normalized and used to construct supervised learning samples, where each model receives the current three-dimensional state vector and predicts the next state.

For comparison, three models are evaluated under the same conditions: a conventional ANN trained using BP, a BPTT-trained SNN, and the proposed UCNN trained using the EAPL rule. The ANN represents the continuous-valued baseline, the SNN represents the purely event-driven baseline, and the UCNN represents the proposed magnitude-phase event-driven learning model. The performance is evaluated using training loss, component-wise RMSE, three-dimensional prediction error, reconstructed attractor geometry, state time histories, and post-training spike activity. Figure 9 summarizes the comparative performance of the ANN, SNN, and proposed UCNN-EAPL models for learning the Lorenz attractor dynamics. This experiment evaluates whether the proposed UCNN framework can generalize beyond the moving-object tracking problem and learn a nonlinear chaotic dynamical system with coupled three-dimensional states. The ANN is used as a continuous-valued baseline, the SNN is used as a purely event-driven baseline, and the UCNN-EAPL model represents the proposed event-driven adaptive phase learning framework. Figure 9(a) shows the training loss histories for the three models. The ANN loss decreases steadily and reaches a moderate final error level, confirming that the feedforward continuous-valued baseline can learn an approximate mapping for the Lorenz system. The SNN training curve also decreases during the early epochs, but it converges to a higher loss level and exhibits a less favorable final performance compared with the ANN and UCNN. This behavior reflects the difficulty of representing smooth chaotic-state evolution using only binary spike-based activity and surrogate-gradient optimization. In contrast, the UCNN-EAPL curve decreases rapidly during the initial epochs and continues to converge toward the lowest final loss. This result indicates that the EAPL rule effectively adapts the UCNN phase dynamics while preserving the continuous-valued information carried by the magnitude channel. Figure 9(b) provides the RMSE comparison for the individual Lorenz state components and for the overall three-dimensional prediction error. The SNN produces the largest RMSE values in all state components, indicating that the purely spiking representation struggles to accurately reconstruct the coupled nonlinear dynamics. The ANN significantly improves over the SNN, but its error remains higher than the UCNN-EAPL model. The proposed UCNN-EAPL achieves the lowest RMSE values for the (x), (y), and (z) components, as well as the smallest overall three-dimensional RMSE. This demonstrates that the magnitude–phase representation is effective not only for visual tracking, but also for nonlinear dynamical-system learning. Figure 9(c) presents the cumulative distribution function of the three-dimensional prediction error. The UCNN-EAPL curve rises most sharply near zero, showing that the majority of UCNN predictions remain within a small error range. The ANN curve also reaches high cumulative probability at relatively low error values, but it is shifted to the right compared with UCNN-EAPL. The SNN curve rises much more slowly and extends over a wider error range, indicating a broader distribution of prediction errors and more frequent large deviations. This distribution-level comparison confirms that UCNN-EAPL provides more consistent predictions across the Lorenz trajectory. Figure 9(d) reports the mean three-dimensional prediction error together with the standard deviation and maximum observed error. The SNN has the largest mean error, largest variability, and highest maximum error, showing that its predictions are more sensitive to the nonlinear and chaotic nature of the Lorenz dynamics. The ANN reduces the error substantially compared with the SNN, but still exhibits larger mean and maximum errors than the proposed model. The UCNN-EAPL produces the lowest mean error, the narrowest variability, and the smallest maximum error among the three models. This confirms that the proposed learning rule improves both accuracy and stability.

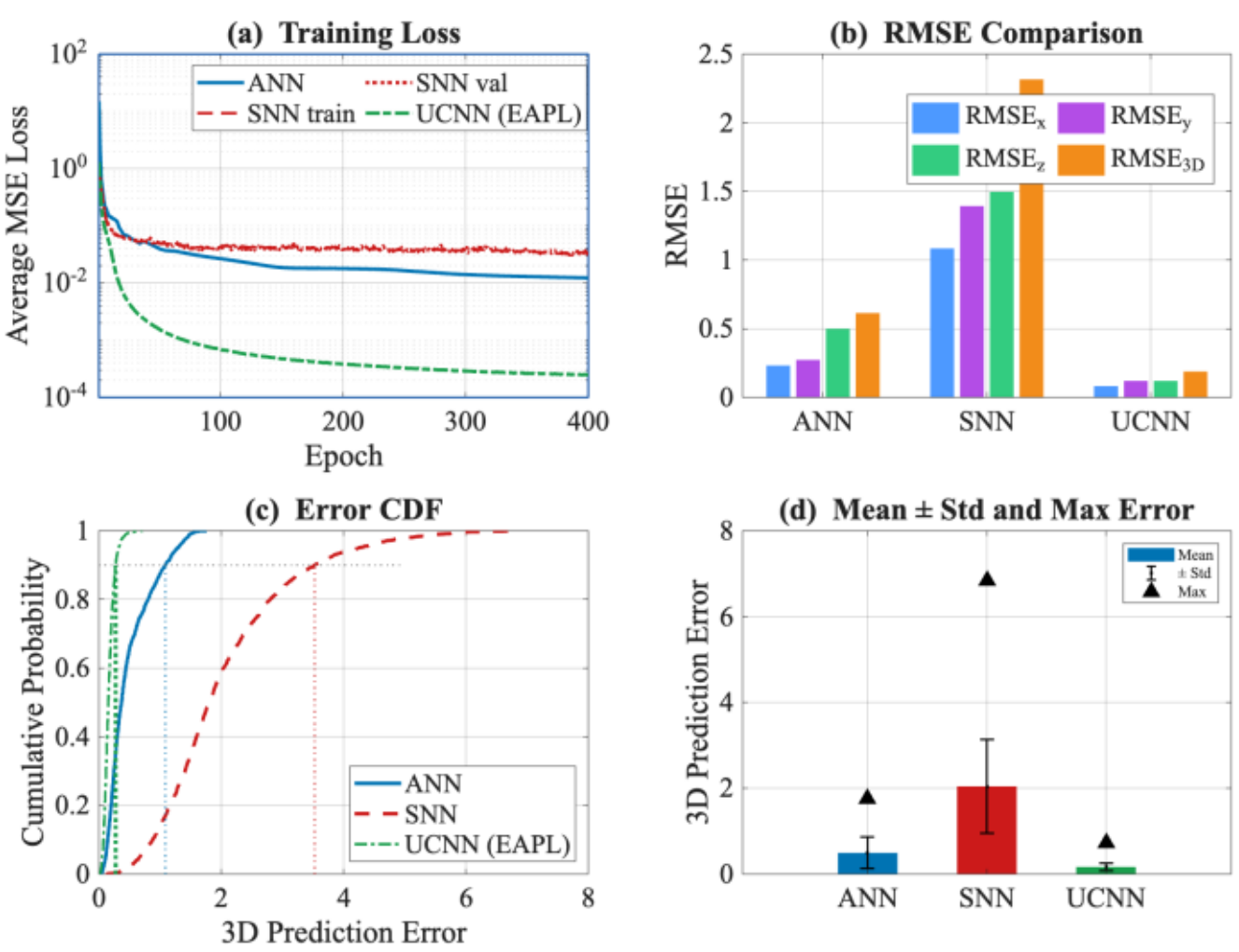

**Figure 9**. Comparative performance evaluation of ANN, SNN, and the proposed UCNN with EAPL on the Lorenz attractor learning task. (a) Training loss histories in terms of average MSE loss. (b) RMSE comparison along the (x), (y), and (z) state components, together with the overall three-dimensional RMSE. (c) Cumulative distribution function of the three-dimensional prediction error. (d) Mean prediction error with standard deviation and maximum observed error. The proposed UCNN-EAPL model achieves the lowest prediction error and the most compact error distribution, demonstrating improved stability and accuracy for nonlinear dynamical-system learning.

Overall, Figure 9 demonstrates that the UCNN-EAPL framework provides a strong advantage for nonlinear spatiotemporal learning. Unlike the ANN, which relies only on continuous-valued static transformations, and unlike the SNN, which relies only on binary event-driven activity, the UCNN combines continuous magnitude encoding with phase-driven temporal dynamics. The EAPL rule further improves this structure by adapting the event-generation dynamics in a computationally efficient manner. Therefore, the Lorenz attractor results support the broader claim that UCNN is not restricted to the object-tracking case study, but can serve as a general neural computational model for learning complex dynamical systems. Figure 10 compares the true Lorenz attractor trajectory with the reconstructed trajectories obtained from the ANN, SNN, and proposed UCNN models. This figure provides a qualitative evaluation of how each architecture captures the geometry of the nonlinear chaotic system in three-dimensional state space. The ANN prediction follows the general structure of the Lorenz attractor and captures the main rotational behavior around the attractor lobes. However, small deviations from the true trajectory can be observed, particularly in regions where the trajectory transitions between lobes or exhibits rapid curvature changes. This behavior is consistent with the quantitative results, where the ANN achieved acceptable accuracy but remained less precise than the UCNN model. The SNN prediction captures the overall attractor shape but exhibits larger deviations from the true trajectory. The reconstructed path shows noticeable distortion and less stable alignment with the ground-truth dynamics, especially during transitions between the two attractor lobes. This confirms that the purely spike-based representation has difficulty preserving the continuous state evolution of the Lorenz system, leading to higher prediction error and larger variability. In contrast, the proposed UCNN model closely overlaps with the true Lorenz trajectory over the full prediction interval. The reconstructed trajectory preserves both the local rotational structure and the global attractor geometry with minimal visible deviation. This result demonstrates that the UCNN magnitude-phase representation is well suited for nonlinear dynamical-system learning. The magnitude channel supports continuous-valued state representation, while the phase channel provides event-driven temporal evolution. Together with the EAPL learning rule, this enables the UCNN to accurately reproduce the Lorenz attractor while maintaining an event-driven computational structure.

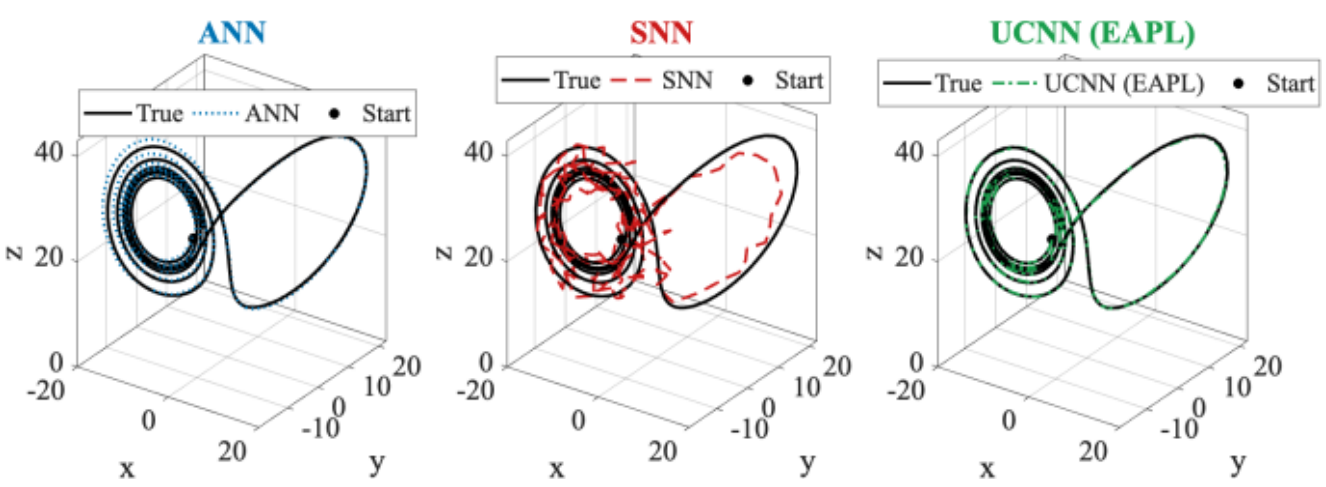

**Figure 10**. Three-dimensional Lorenz attractor reconstruction obtained using ANN, SNN, and the proposed UCNN model. The black curve represents the true trajectory, and the colored curves show the corresponding model predictions, with the starting point marked by a black dot

Figure 11 shows the smoothed three-dimensional prediction-error histories obtained from the ANN, SNN, and proposed UCNN models over the Lorenz attractor test sequences. This figure provides a time-indexed comparison of prediction stability and complements the RMSE, CDF, and reconstructed-attractor results presented earlier. The ANN maintains a moderate prediction error across the sequence. Although its error remains significantly lower than that of the SNN, the ANN curve exhibits repeated oscillatory peaks, indicating sensitivity to specific regions of the Lorenz trajectory. These peaks are expected in nonlinear chaotic systems, where small local prediction deviations can grow rapidly during state transitions or high-curvature regions of the attractor. The SNN exhibits the largest error throughout the sequence, with several pronounced peaks. This behavior indicates that the purely spike-based model has difficulty preserving the continuous state evolution of the Lorenz system. Since the Lorenz attractor requires accurate reconstruction of coupled continuous dynamics, relying only on binary spike events leads to higher prediction variability and larger transient deviations. In contrast, the proposed UCNN maintains the lowest and most stable prediction error over nearly the entire sequence. The UCNN curve remains close to zero and shows only small fluctuations compared with both ANN and SNN. This result confirms that the magnitude-phase formulation improves nonlinear dynamical-system learning by combining continuous-valued state representation with event-driven temporal evolution. Therefore, the smoothed error history demonstrates that UCNN provides more accurate and stable Lorenz prediction than both baseline models.

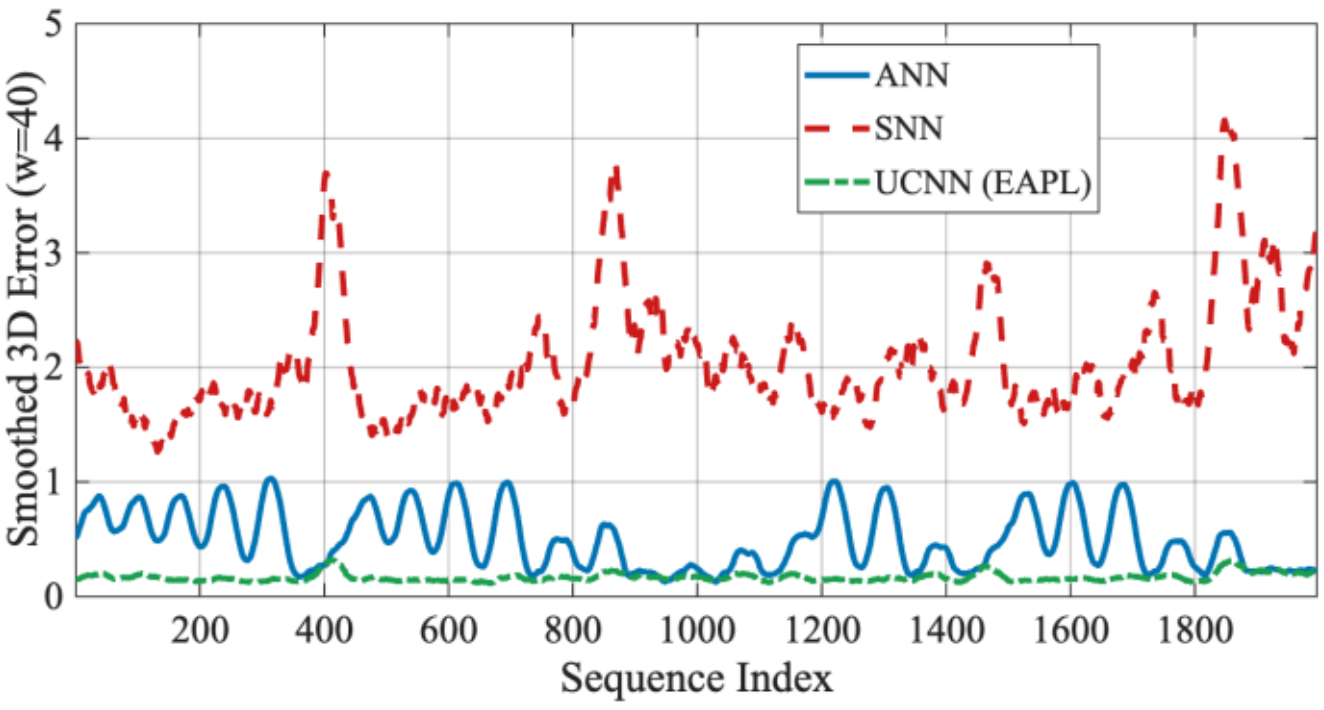

**Figure 11**. Smoothed three-dimensional prediction-error histories for ANN, SNN, and the proposed UCNN on the Lorenz attractor learning task.

Figure 12 compares the true Lorenz state trajectories with the predictions obtained from ANN, SNN, and the proposed UCNN over the test interval. The ANN follows the general trend of all three states with small deviations, while the SNN shows more visible fluctuations and local mismatches, especially near sharp transitions and high-curvature regions. In contrast, the UCNN prediction remains closely aligned with the true trajectories for x, y, and z. This confirms that the proposed magnitude-phase representation can accurately capture the coupled temporal evolution of the Lorenz system and provides more stable state prediction than the purely spiking baseline.

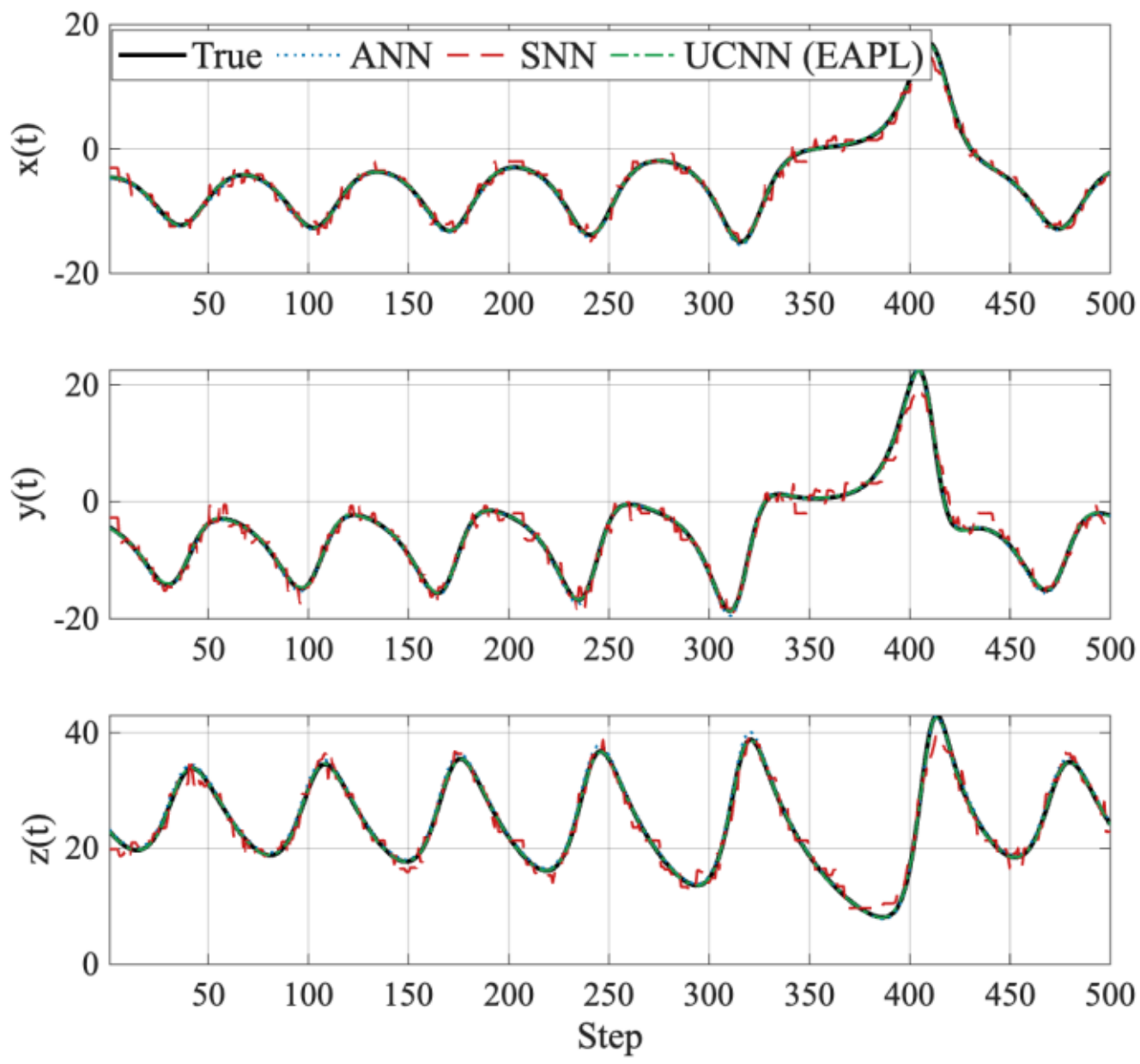

**Figure 12**. Time-history comparison of the true and predicted Lorenz states using ANN, SNN, and the proposed UCNN

Figure 13 compares the post-training event activity of the SNN and UCNN for the Lorenz attractor case study. The SNN exhibits irregular and input-dependent spiking patterns in both layers, reflecting the difficulty of encoding continuous chaotic dynamics using binary spike events alone. In contrast, the UCNN produces more regular and structured activity across both layers. This behavior indicates that the phase dynamics of the UCNN provide a stable temporal representation, while the magnitude channel preserves continuous-valued information. Therefore, the observed spike activity supports the improved prediction accuracy and stability of UCNN in learning the Lorenz attractor.

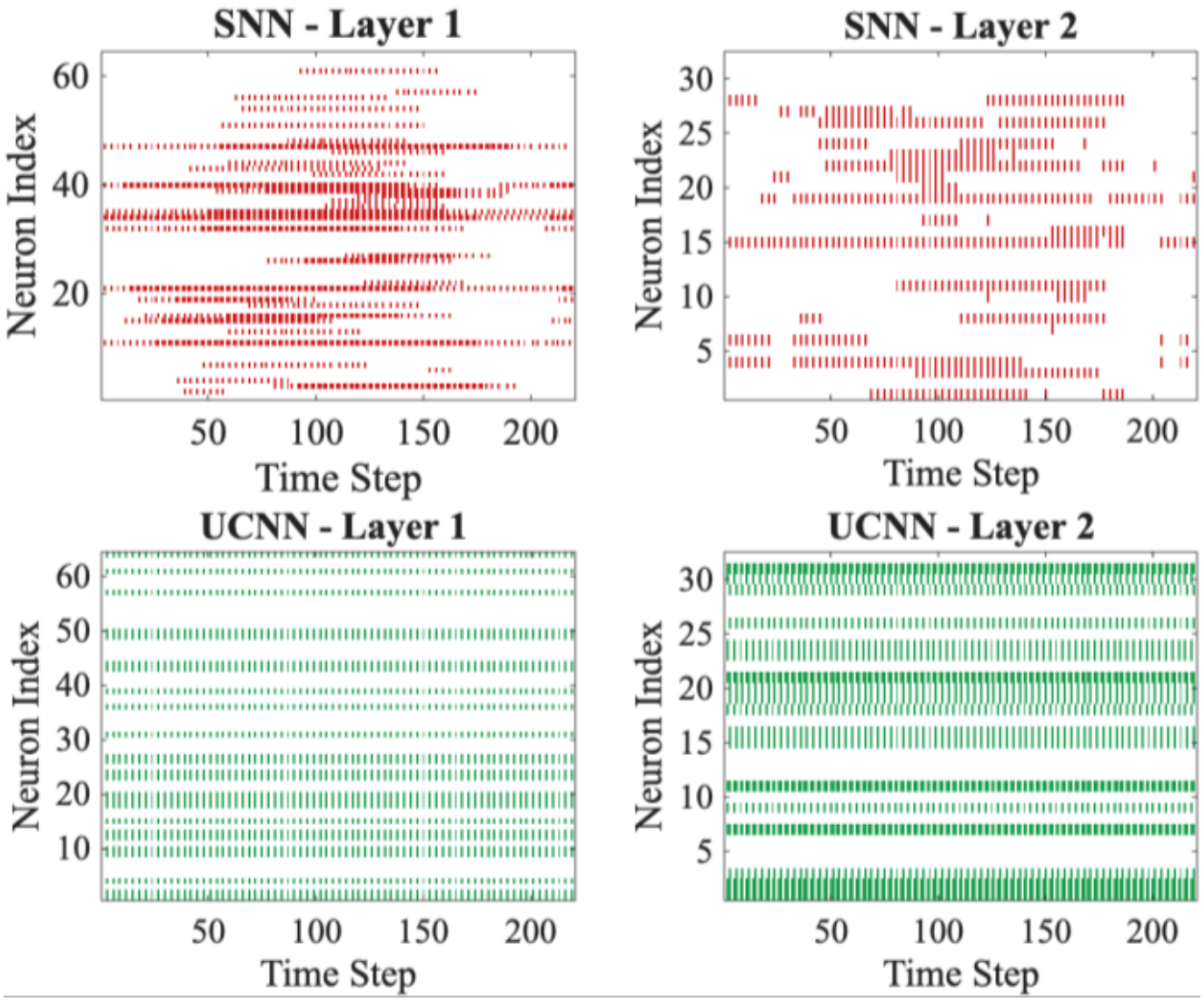

**Figure 13**. Post-training spike activity comparison between the SNN and the proposed UCNN for the Lorenz attractor learning task across two hidden layers.

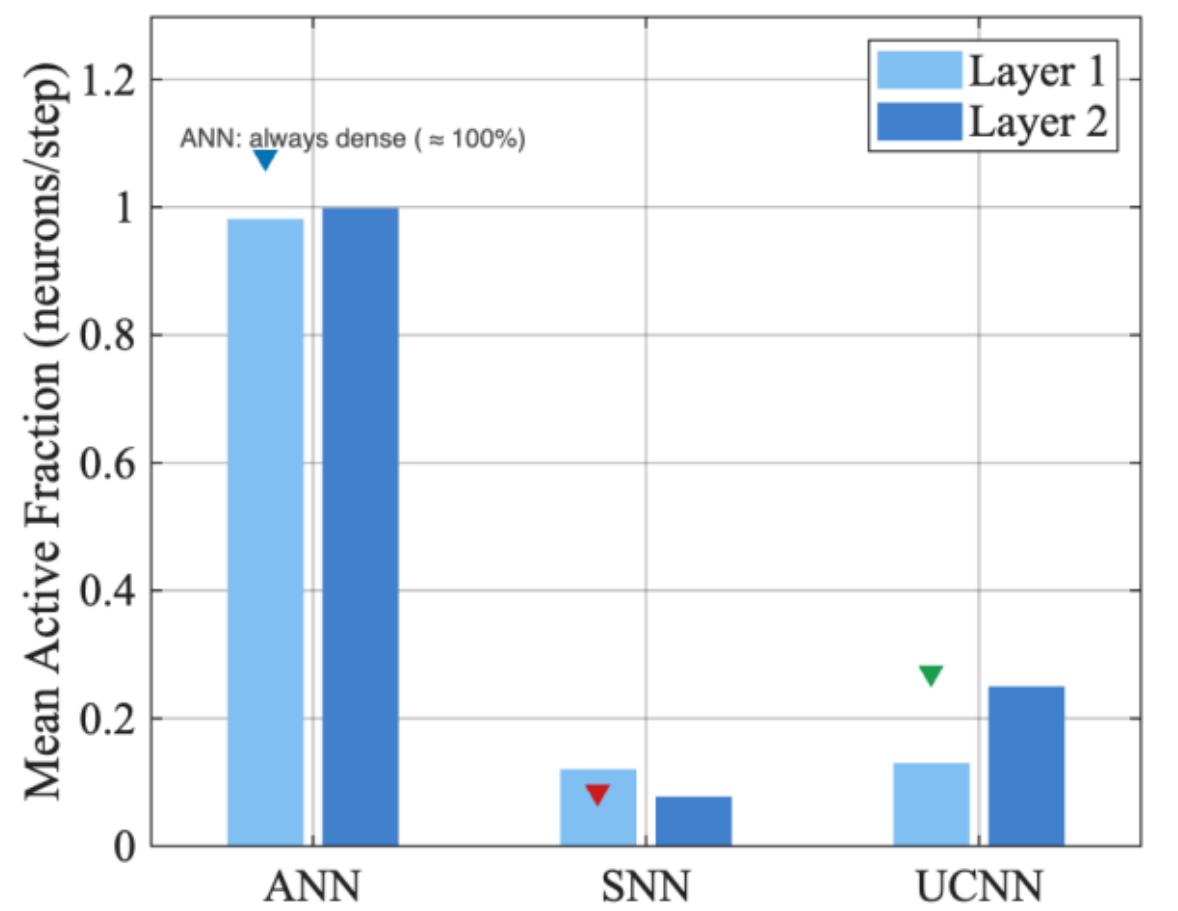

**Figure 14**. Mean active fraction of neurons per step for ANN, SNN, and UCNN in the Lorenz attractor learning task.

Figure 14 compares the mean active fraction of neurons per step across the two hidden layers for the ANN, SNN, and UCNN models. The ANN remains almost fully dense in both layers, since all neurons participate in computation at each step. The SNN shows the lowest activity due to its sparse binary spike-based operation. The proposed UCNN also operates sparsely, with activity much lower than the ANN while remaining slightly higher than the SNN in the second layer. This confirms that UCNN preserves event-driven sparsity while maintaining richer magnitude–phase information flow for accurate Lorenz attractor prediction.

## 5 Conclusion

This paper introduces the unified complex-valued neural network (UCNN), bridging the continuous precision of ANNs with the sparse, energy-efficient temporal processing of SNNs. By leveraging an asymmetric complex-valued state, the UCNN uses magnitude to encode graded signal strength and phase to govern intrinsic temporal event generation, effectively eliminating the architectural overhead of traditional hybrid neural networks. To efficiently optimize this architecture end-to-end, we developed a BP-BPTT framework and the event-driven adaptive phase learning (EAPL) rule, which minimizes computational burden by circumventing explicit temporal unrolling via adjoint-style recursion. Evaluated on spatiotemporal object tracking and chaotic Lorenz attractor prediction, the UCNN preserved sparse event-driven computation while achieving an acceptable tracking error reduction compared to purely spiking baselines. These results establish the UCNN as a highly robust and scalable framework for neuromorphic computing and edge-AI.

## 6 Conflict of Interest

The authors declare that they have no conflicts of interest related to this research. The study was conducted in an objective and unbiased manner, and the results presented herein are based on rigorous analysis and interpretation. The authors have no financial or personal relationships with individuals or organizations that could potentially bias the findings or influence the conclusions of this study.

## References

[1] Y. LeCun, Y. Bengio, and G. Hinton, "Deep learning," Nature, vol. 521, pp. 436–444, 2015.

[2] L. Deng et al., "Rethinking the performance comparison between SNNS and ANNS," Neural Networks, vol. 121, pp. 294–307, 2020.

[3] K. Roy, A. Jaiswal, and P. Panda, "Towards spike-based machine intelligence with neuromorphic computing," Nature, vol. 575, pp. 607–617, 2019.

[4] M. Davies et al., "Advancing neuromorphic computing with Loihi: A survey of results and outlook," Proceedings of the IEEE, vol. 109, no. 5, pp. 911–934, 2021.

[5] C. D. Schuman et al., "Opportunities for neuromorphic computing algorithms and applications," Nature Computational Science, vol. 2, pp. 10–19, 2022.

[6] D. Kudithipudi et al., "Neuromorphic computing at scale," Nature, vol. 637, pp. 801–812, 2025.

[7] J. Pei et al., "Towards artificial general intelligence with hybrid Tianjic chip architecture," Nature, vol. 573, pp. 106–111, 2019.

[8] G. Gallego et al., "Event-based vision: A survey," IEEE Transactions on Pattern Analysis and Machine Intelligence, vol. 44, no. 1, pp. 154–180, 2022.

[9] Y. Zhou et al., "Computational event-driven vision sensors for in-sensor spiking neural networks," Nature Electronics, vol. 6, pp. 870–878, 2023.

[10] Y. Wang et al., "A biologically inspired artificial neuron with intrinsic plasticity based on monolayer molybdenum disulfide," Nature Electronics, vol. 8, pp. 680–688, 2025.

[11] R. Zhao et al., "A framework for the general design and computation of hybrid neural networks," Nature Communications, vol. 13, p. 3427, 2022.

[12] J. K. Eshraghian et al., "Training spiking neural networks using lessons from deep learning," Proceedings of the IEEE, vol. 111, no. 9, pp. 1016–1054, 2023.

[13] W. Fang et al., "SpikingJelly: An open-source machine learning infrastructure platform for spike-based intelligence," Science Advances, vol. 9, eadi1480, 2023.

[14] H. Zheng et al., "Temporal dendritic heterogeneity incorporated with spiking neural networks for learning multi-timescale dynamics," Nature Communications, vol. 15, 277, 2024.

[15] A. Stanojevic et al., "High-performance deep spiking neural networks with 0.3 spikes per neuron," Nature Communications, vol. 15, 6793, 2024.

[16] J. Ding et al., "Neuromorphic computing paradigms enhance robustness through spiking neural networks," Nature Communications, vol. 16, 10175, 2025.

[17] Y. Yan et al., "Efficient and robust temporal processing with neural oscillations modulated spiking neural networks," Nature Communications, 2025.

[18] S. D. Pokala et al., "A frugal spiking neural network for unsupervised multivariate temporal pattern classification and multichannel spike sorting," Nature Communications, vol. 16, 9218, 2025.

[19] T. Shi et al., "Fully memristive spiking neural network for energy-efficient graph learning," Science Advances, 2025.

[20] Y. Guo et al., "Direct learning-based deep spiking neural networks: A review," Frontiers in Neuroscience, vol. 17, 1209795, 2023.

[21] C. Yu et al., "Advancing training efficiency of deep spiking neural networks through rate-based backpropagation," Advances in Neural Information Processing Systems, 2024.

[22] C.-Y. Lee et al., "Complex-valued neural networks: A comprehensive survey," IEEE/CAA Journal of Automatica Sinica, vol. 9, no. 8, pp. 1406–1426, 2022.

[23] T. Wu, A. Scaglione, and D. Arnold, "Complex-value spatiotemporal graph convolutional neural networks and applications to electric power systems," IEEE Transactions on Smart Grid, 2023.

[24] J. Zhang et al., "Event-based sensor fusion and application on odometry: A survey," 2024.

[25] H. Wang et al., "Event camera meets mobile embodied perception: Abstraction, algorithm, acceleration, application," ACM Computing Surveys, 2026.

[26] R. Ahmadvand, S. S. Sharif, and Y. M. Banad, "Neuromorphic robust framework for integrated estimation and control in dynamical systems using spiking neural networks," Scientific Reports, 2025.

[27] R. Ahmadvand et al., "Cloud-edge event-driven control using spiking neural networks and local plasticity rules," Neuromorphic Computing and Engineering, 2024.

[28] W. He et al., "Spiking graph neural networks on Riemannian manifolds," Advances in Neural Information Processing Systems, 2024.

[29] Y. Chen et al., "Fully memristive spiking neural network for energy-efficient graph learning," Science Advances, 2025.

[30] R. Ahmadvand, S. S. Sharif, and Y. M. Banad, "A unified phase-native computational principle governs hippocampal spike timing and neural coding," arXiv preprint arXiv:2603.19690, 2026.